\documentclass[12pt]{article}

\usepackage{palatino}
\usepackage{color}
\usepackage{graphicx}

\usepackage{tikz-qtree}
\usepackage{amsfonts}
\usepackage{amsmath}
\usepackage{hyperref}
\hypersetup{colorlinks=true,citecolor=blue, linkcolor=blue, urlcolor=blue}
\usepackage{subfigure}

\usepackage{fullpage}

\definecolor{Red}{RGB}{255,0,0}
\definecolor{Green}{RGB}{0,255,0}
\definecolor{Blue}{RGB}{0,0,255}
\definecolor{LightBlue}{RGB}{150,150,255}

\newcommand{\fj}[1]{}
\newcommand{\fjurgent}[1]{}
\newcommand{\fjdone}[1]{}

\newcommand{\tableref}[1]{Table~\ref{#1}}
\newcommand{\figref}[1]{Figure~\ref{#1}}

\title{Human languages order information efficiently}
\author
{Daniel Gildea,$^{1}$ and T. Florian Jaeger$^{1,2,3}$\\
\\
\normalsize{$^{1}$Department of Computer Science}\\
\normalsize{$^{2}$Department of Brain and Cognitive Sciences}\\
\normalsize{$^{3}$Department of Linguistics}\\
\normalsize{University of Rochester}\\
\normalsize{Rochester, NY 14627, USA}\\
}
\date{}

\begin{document} 
\maketitle 

\begin{abstract}
  Most languages use the relative order between words to encode
  meaning relations. Languages differ, however, in what orders
  they use and how these orders are mapped onto different
  meanings. We test the hypothesis that --despite these
  differences-- human languages might constitute different
  `solutions' to common pressures of language use. Using Monte
  Carlo simulations over data from five languages, we find that
  their word orders are efficient for processing in terms of both 
  dependency length and local lexical probability. This suggests
  that biases originating in how the brain understands language
  strongly constrain how human languages change over generations.
\end{abstract}

\section{Introduction}
\label{sec:overview}
We test the hypothesis that language change is subject to small but persistent biases that result, on average, in languages that are easier to process. Biases for grammars with higher processing efficiency could be the direct result of abstract learning biases \cite{Slobin1975, Fedzechkina2012} or they could result from the pressures of language use \cite{Bybee2001, hawkins94, Hawkins2004}, such as preferences that have been hypothesized to operate during language production \cite{Gibson2013, Jaeger2010, Lindblom1990} or biases originating in comprehension \cite{Guy1996, Ohala1988, Pierrehumbert2002, Wedel2006}. 


If language change is indeed subject to biases towards languages with higher processing efficiency and if these biases are sufficiently strong, these biases should over accumulate over historical time, leading natural languages that have existed for sufficiently long to have higher than expected processing efficiency. This is the primary hypothesis we set out to test. Some evidence suggests that the sound structure and lexicon of natural languages exhibit properties that are expected under this hypothesis \cite{Graff2009, Manin2006, Piantadosi2011, Piantadosi2012, Zipf1949}. At those levels of linguistic organization, studies over the last couple of years have also provided more direct correlational evidence that language change is affected by processing \cite{Wedel2013a}. Miniature language learning experiments have documented similar biases during language acquisition and that these biases can accumulate over generations of learners \cite{Kirby2008}. 

The level of linguistic organization that has remained elusive with regard to this question, however, is also arguably the one that is the one that makes human languages most unique compared to all other animal communication systems: syntax --or some aspects of syntax (recursion)-- give human languages infinite expressivity with finite means \cite{Humboldt1972, Nowak2000} and it is syntax that has been taken to be the defining property of human languages (e.g., \cite{Hauser2002}; but see \cite{Pinker2005}). Whether at least some properties of the syntactic systems of languages can be derived from the fact that languages need to be processed continues to be a heatedly debated question (for recent high impact reviews, see \cite{Fedzechkina2012, Kirby2007, Kirby2008, Pinker2005}).
One reason why this question has not been directly addressed, as we detail below, is that until very recently it has been {\em impossible} to directly test whether the syntax of natural languages tends to facilitate processing efficiency. Here we present the results of several large-scale computational simulations that address these questions. For the purpose of presentation, we group these simulations into Studies 1 and~2.

Study 1 tests and finds confirmed the hypothesis that natural languages have word orders that makes them easier to process than expected by chance. From this is does not follow that natural languages have {\em optimal} or even close to optimal processing efficiency. Processing efficiency is presumably just one of several factors that might bias language change (the ease of acquisition of a grammar being another constraint). 
Still, if biases towards efficient processing are among the most influential factors influencing language change, we would expect human languages to have word orders that are pretty close to optimal in terms of processing efficiency. This hypothesis is tested in Study 2. Taken together, Studies 1 and 2 suggest that language processing exhibits a surprisingly strong bias on language change.

The hypothesis we test is one that has long intrigued language researchers. The pressures inherent to language processing have long been assumed to shape languages over time, including not only phonology and the lexicon \cite{Kohler1991, Lindblom1990, Ohala1988, Hume2013, Zipf1949}, but also syntactic structure \cite{Bybee2001, Bates1982, Bates1987, hawkins94, Slobin1975}. However, until relatively recently it has virtually been impossible to obtain reliable estimates of the processing efficiency of a language. Imagine one was to obtain such estimates experimentally (e.g., by obtaining estimates of the word-by-word processing times a native speaker of that language experiences while reading sentences from that language). A reliable estimate of the processing efficiency of an entire language would require reading data for a representative sample of the language. Ideally, this sample would be representative in terms of its lexical and grammatical distributions --i.e., it should contain both low and high frequency words, more and less complex syntactic structures, and so on. Further reliable estimates would require that individual differences in, for instance, reading abilities are averaged out. In short, hundreds of readers would likely have to read thousands of sentences. This alone is a daunting task. In one of the two most commonly used methods to obtain word-by-word reading time estimates (self-paced reading), it takes between .5-1 hour to obtain reading times for 100 sentences. So, to obtain data from 100 readers on, say 1000 sentences from a language --which would still not be a lot of sentences--, we would require about 500-1000 participant hours.

However, by far the biggest challenge lies in establishing a chance-level against which to compare the processing efficiency of a language. This requires estimates of processing efficiency from a large set of randomized {\em variants} of a language (see below). This further increases the required experimental data by several orders of magnitude. Assessing the processing efficiency of a language based on human data is thus prohibitively expensive and time-consuming. The {\em smallest} study we present below would correspond to 500,000 participant hours. At New York State minimum wage (as of 12/31/2014), this approach of assessing processing efficiency would cost over 4 million US Dollars per language, for a total of 20 million Dollars for the five languages we examine. It would also arguably provide an utterly anti-conservative estimate of chance (to say the least): without extensive training on the new language variant, participants would experience massive interference from their native language, making it {\em appear} as if human languages are highly efficient simply because it is the one that participants are familiar with (it takes most learners of a language years to have approximately native-like processing speeds).

Here, we take an alternative approach. We take advantage of advances in computational psycholinguistics, natural language processing, and the availability of large linguistic databases. Rather than to obtain estimates of processing efficiency from human readers, we automatically estimate the processing efficiency of a language from large linguistically annotated collections of text (syntactically annotated corpora). This is now possible, because psycholinguistic research has identified grammar-dependent measures of processing efficiency. Here, we focus on two properties that are known to affect word-by-word processing times: a word's Shannon information in context (i.e., its {\em surprisal} \cite{Hale2001, Levy2008}) and the length of the dependencies that are integrated at the word ({\em dependency length}, \cite{Gibson1998, Gibson2000}).

We describe and further motivate these two measures in more detail in below. For now, it suffices to say that processing difficulty (as assessed through, e.g.,  per-word reading times) is positively correlated with surprisal and dependency length. If a bias for processing efficiency affects the development of languages over time, it is thus expected that natural languages have lower average surprisal and shorter average dependency lengths than expected by chance. 

We test these predictions against data from five languages: Arabic (Modern Standard), Czech, English (American), German, and Mandarin Chinese. These five languages were chosen for two reasons. First, we aimed for representative linguistic coverage. Languages often share linguistic properties simply because they are historically related or because they have co-existed in geographic proximity over long periods of time, with the ensuing language contact leading to lexical and grammatical borrowings. Here, we are interested in testing hypotheses that are assumed to apply universally across {\em all} languages. The less historically and geographically related the languages in our sample are, the more likely any effect found on this sample is to generalize beyond the particular sample to {\em any} language. 

The five languages we investigated represent three major language families (Sino-Tibetan,  Indo-European, and Semitic) and four language subfamilies (Chinese, Balto-Slavic, Germanic, and Arabic). The five languages also differ in a variety of linguistic properties that are known to be relevant to processing difficulty. For example, three of the languages in our sample have dominant Subject-Verb-Object (SVO) order, one of them has dominant VSO order (Arabic), and one has no dominant word order (German). The languages also differ in whether they productively use morphological means to mark grammatical relations, such as using case (Arabic, Czech, German), or not (English, Mandarin). As a third and final example, the languages differ in whether and how they express certain arguments to the verb. For example, pronominal elements in subject position (e.g., {\it I, you, he}) can optionally be omitted in Mandarin, are realized as suffixes on the verb in Czech, but are more or less obligatorily realized as separate words in English and German. Any of these properties could theoretically affect the measures we assess in our studies. 

Second, as we describe next, sufficiently large electronic corpora with the necessary linguistic annotations are now available for these languages. Corpus size is critical for our purpose. The reliability of the estimates we derive below depends on the number of words and sentences in the corpus. For example, the accuracy and reliability of the processing efficiency estimates described below increases with the number of words in a corpus. The corpora we employ in our studies are the largest available corpora for the five languages with the required linguistic annotation. The methods we use to obtain surprisal and dependency length estimates further increase robustness of estimates.

\section{Data}
\label{sec:data}

The data for all languages comes from newspaper corpora. For English, we also had access to a corpus of conversational speech data with the required annotations. An overview of the corpora is provided in \tableref{t:corpus}.

\begin{table}
\begin{tabular}{l|r|rrrr}
	& sentences & \multicolumn{4}{c}{sentence length} \\
	& & mean & std dev & min & max \\
\hline
Arabic (Modern Standard)	&  6,776 & 35.4 & (26.9) & 1 & 387\\
Czech	& 72,703 & 14.8 & (9.6) & 1 & 166\\
English (American, written) & 39,832 & 20.9 & (10.1) & 1 & 122\\
English (American, spoken) & 17,968 & 7.9 & (8.2) & 1 & 92\\
German	& 45,422 & 15.5 & (9.6) & 1 & 115\\
Mandarin Chinese	& 28,289 & 23.8 & (16.4) & 1 & 212\\
\end{tabular}
\caption{Overview of corpora used in the current studies}\label{t:corpus}
\end{table}

Specifically, the Arabic data consists of 6776 sentences from the Penn Arabic Treebank, in the dependency representation of the Prague Arabic Dependency Treebank
version 1 \cite{hajic2004prague}.
The Czech data consists of 72,703 sentences from the
Prague Dependency Treebank version 1 \cite{PDT},
as used in the CoNLL 2006 dependency parsing
evaluation \cite{buchholz2006conll}.
The English data comes from two sources. For written data, we use the 39,832 sentences from the
Wall Street Journal portion of the Penn Treebank version 3 \cite{treebank}.
For spoken data, we use 17,968 sentences from the Switchboard corpus of spoken English \cite{Godfrey92}.
The German data consists of 45,422 sentences from the 
TIGER corpus \cite{TIGER}, which primarily consists of articles from the German newspaper ``Frankfurter Rundschau''.  
Finally, the Mandarin Chinese data consists of 28,289 sentences from 
the Penn Chinese Treebank version 6.0 \cite{ctb}. This includes newswire from Xinhua News Agency, articles from Sinorama Magazine, news from the website of the Hong Kong Special Administrative Region and transcripts from various broadcast news programs.

\begin{figure}
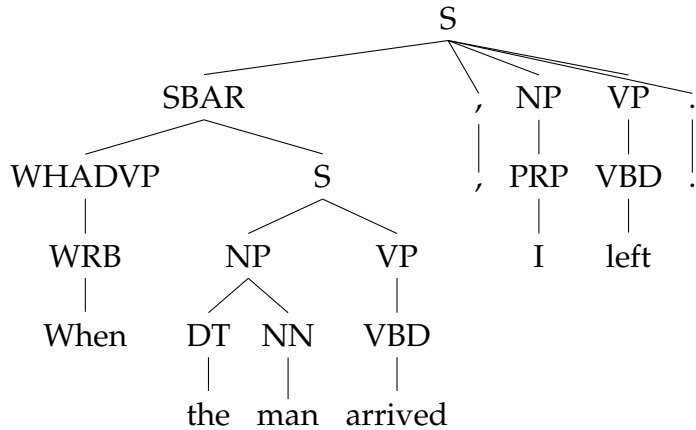


\Tree[.S [.SBAR [.WHADVP [.WRB When ]  ]  [.S [.NP [.DT the ]  [.NN man ]  ]  [.VP [.VBD arrived ]  ]  ]  ]  [., , ]  [.NP [.PRP I ]  ]  [.VP [.VBD left ]  ]  [.. . ]  ]  
\caption{Example syntax tree}\label{fig:tree}
\end{figure}

 All corpora consist of sentences that have been 
manually annotated with the syntactic structure of each
sentence.  The annotations specify a syntactic structure for
each sentence. The annotation types differed somewhat between languages. An example, from the English corpus is shown in \figref{fig:tree}. 

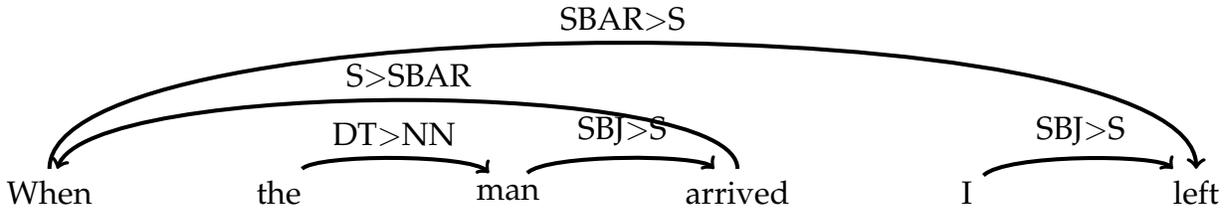
\begin{figure}
\begin{center}
\resizebox{\linewidth}{!}{
\begin{tikzpicture}
\draw (0,0) node(1) {When};
\draw (3,0) node(2) {the};
\draw (6,0) node(3) {man};
\draw (9,0) node(4) {arrived};
\draw (12,0) node(5) {I};
\draw (15,0) node(6) {left};
\draw[line width=1.5pt,->] (2) .. controls +(0.5,0.5) and +(-0.5,0.5) ..node[above]{DT$>$NN} (3);
\draw[line width=1.5pt,->] (3) .. controls +(0.5,0.5) and +(-0.5,0.5) ..node[above]{SBJ$>$S} (4);
\draw[line width=1.5pt,->] (4) .. controls +(0,1.5) and +(0.5,1.5) ..node[above]{S$>$SBAR} (1);
\draw[line width=1.5pt,->] (1) .. controls +(0,2.5) and +(0,2.5) ..node[above]{SBAR$>$S} (6);
\draw[line width=1.5pt,->] (5) .. controls +(0.5,0.5) and +(-0.5,0.5) ..node[above]{SBJ$>$S} (6);
\end{tikzpicture}}
\end{center}
\caption{Dependency structure, converted from the syntactic tree in \figref{fig:tree}}\label{fig:dep}
\label{graph}
\end{figure}

We automatically converted the different syntactic annotations into a 
dependency representation, as shown in \figref{fig:dep}. We use the dependency 
representation because dependency length
has been shown to be an important variable affecting 
human language processing (see below). The dependency representation
is a directed graph specifying, for each word in the 
sentence, the {\em head word} (or `sender' \cite{Ferrer2004}) that it modifies.  For 
example, subjects and direct objects modify the main verb
of the clause; determiners, adjectives, and relative
pronouns typically modify nouns;
prepositions can modify nouns or verbs;
prepositions are modified by the object nouns; and so on.
We convert trees to
dependency representations using a set of rules 
which specify which child of each node in the tree
is the head child, i.e., the main component of the phrase \cite{Magerman94,Collins99}.
Recursively choosing a head child for each node from
the top down, we find a head word for each node in the
tree.  At each node in the tree, dependency relations
are created indicating that the head word of the head child is
modified by the head word of each other child.
The dependency representation tends to be robust to the 
details of the syntactic annotation schemes used 
by various corpora.  

For English, German, and Chinese, we extracted dependencies
from constituent representations, converting the representation
of Figure~\ref{fig:tree} to that of Figure~\ref{fig:dep}.
Specifically, we extract dependencies using the head-finding rules
of Collins \cite{Collins99}.  Our dependency types consist of
pairs of syntactic categories, with one element representing
the category of the maximal projection of the head, and
one representing the category of the maximal projection of the 
modifier. Additionally, we include a special subject type
in order to differentiate verb subjects and direct objects,
by using the ``SBJ'' function tag in the Penn treebank annotation (see \figref{fig:dep}).

For Czech and Arabic, our data was originally annotated in 
a dependency representation.  We take advantage of 
relation labels provided, which included relations such subject, object,
attribute, and so on.  Our dependency types consist of
both the relation of a word and the relation of its parent,
in order to allow us to distinguish between, for example,
an attribute relation in a subject noun phase and an attribute
relation modfifying a verb in a relative clause.

\section{Estimating the Processing Efficiency of Languages}\label{sec:processibility}

As outlined in the introduction, we focus on two measures of processing efficiency that have received broad empirical support: {\em surprisal} \cite{Hale2001, Levy2008} and {\em dependency length}, \cite{Gibson1998, Gibson2000}.  

The surprisal of a word is identical to its Shannon information (in bits) in context, which is defined as the logarithm (to base 2) of the inverse of its probability in context. 
\begin{eqnarray}
I(w) &=& \log_2{\frac{1}{p(w|context)}} \\
& = & -\log_2{p(w|context)}
\end{eqnarray}

A word's surprisal (conditioned on all relevant preceding context) has been shown to be identical to the relative entropy (or Kullback-Leibler divergence) between the distribution over all possible parses prior to the word and the distribution over all possible parses after processing the word \cite{Levy2005}. Surprisal can thus be understood as a measure of the amount of syntactic belief-updating that is associated with processing the word. Crucially, a word's surprisal has been found to be a good predictor of its reading times in context \cite{Boston2008, Demberg2008, Frank2011, McDonald2003, Smith2013}. For example, in a large-scale reading experiment, Smith and Levy \cite{Smith2013}  found that per-word reading times were linear in the word's surprisal. This relation held over six orders of magnitude in the probability, from almost perfectly predictable instances of words to  barely predictable instances ($1 \geq p(word|context) \geq .000001$). Surprisal has also been found to be reflected in neural responses. \fj{(for a review, see \cite{Kuperberg2015}).}

Dependency length, too, has been found to affect processing difficulty, with longer dependencies leading to longer reading times at their integration point. Consider the word {\em left} in the example in \figref{fig:dep}. Two dependencies end --and are thus assumed to be integrated-- at the word {\em left}. One is the dependency between the verb {\em left} and its subject ({\em I}). This dependency is local. The other dependency is between the verb and its temporal modifier ({\em when the man arrived}). This dependency is non-local. Psycholinguistic research has found that non-local dependencies tend to cause processing difficulty (\cite{Gibson1998, Gibson2000, Grodner2005}; though see \cite{Lewis2006, Vasishth2005} for discussion). 
There is also evidence that cross-linguistically speakers prefer shorter dependencies over longer ones when their language provides them with two ways of encoding a message (e.g., for Basque \cite{Ros2015}; English: \cite{Arnold2000, Arnold2004, Lohse2004}; Japanese \cite{Yamashita2001}; Korean \cite{Choi2007}; for reviews and discussion, see \cite{Hawkins2014, JaegerNorcliffe2009}). 

\begin{figure}
\begin{center}
\includegraphics[width=4.5in]{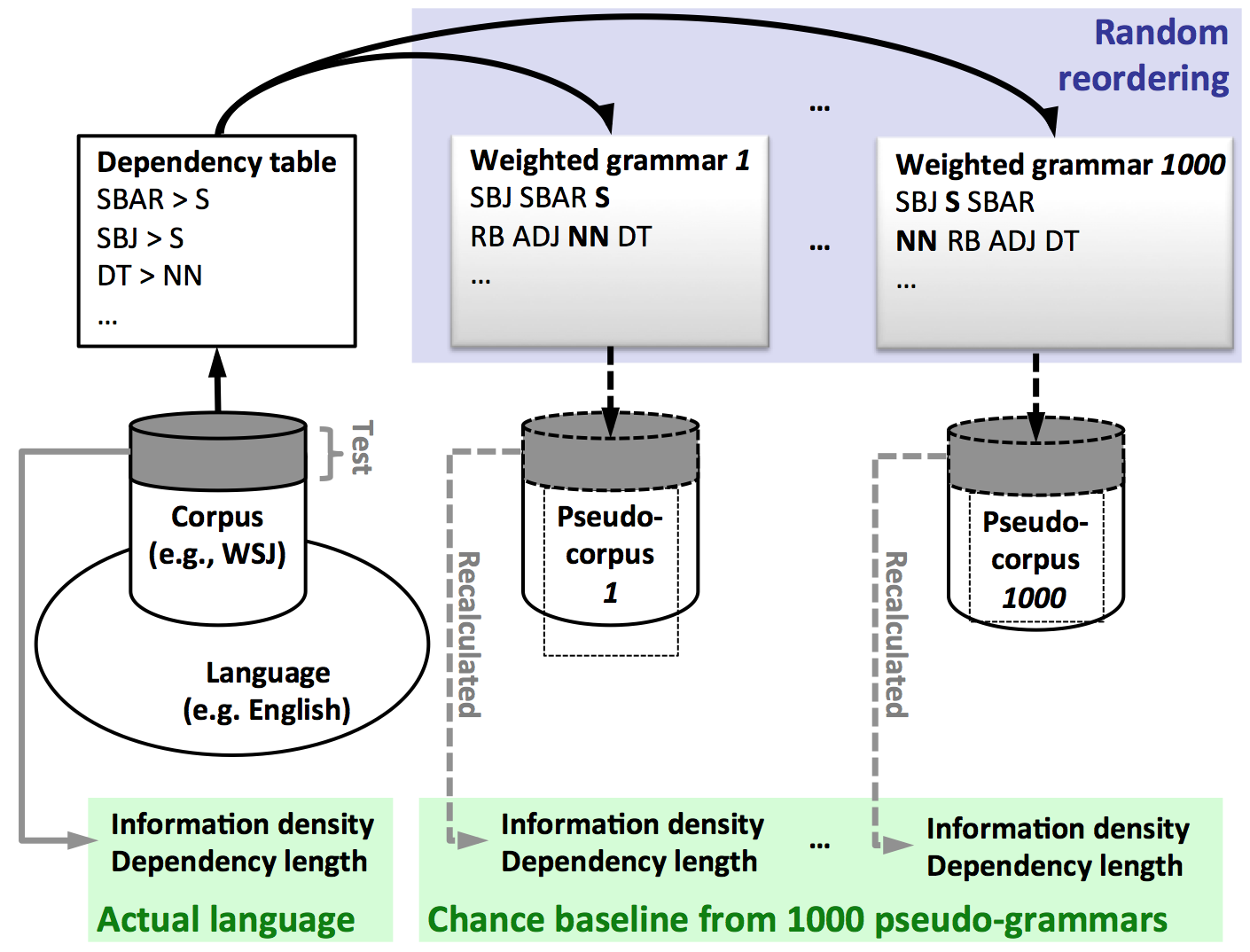} 
\caption{Overview of procedure used to compare the processing efficiency of natural languages (measured in terms of their average information density and dependency length) against the baseline efficiency expected by chance. Weighted grammars are the random reorderings of sets of dependencies (see text).}\label{fig:procedure}
\end{center}
\end{figure}

Here, we estimate these two measures for entire languages. That is, unlike in psycholinguistic work, which has focused on the word-by-word effects of surprisal and dependency length on language processing, we are estimating surprisal and dependency length at the system level. To us, these measures are of interest because they provide an estimate of the average processing difficulty a native speaker of a language experiences while processing that language. This allows us to test whether natural languages have lower average surprisal and shorter average dependency lengths than expected by chance. An overview of the procedure is given in \figref{fig:procedure}.

There are other factors that are known to contribute to processing efficiency. For example, among the primary contributors to word-by-word processing are {\em lexical} properties. To name just a few of these properties, a word's length, frequency, neighborhood density, part-of-speech, and morphological structure are all correlated with the average time it takes to comprehend or produce that word \cite{Baayen2006, Balota2004, Boston2008, DelPradoMartin2004a, Demberg2008, Luce1998, Magnuson2007b}. As expected under the general hypothesis tested here, several studies have found the lexicon of languages to exhibit properties that are consistent with the hypothesis that processing efficiency over time shapes the phonology of words \cite{Bybee1999, CohenPriva2008, Graff2009, Manin2006, Piantadosi2011, Piantadosi2012, Zipf1949}. Here, however, we are interested in a {\em grammatical} property of languages --specifically, word order-- and how it affects processing efficiency. It is grammatical properties that would differ between grammatical systems, thus allowing processing preferences to affect `selection' of these properties over time. The approach we present below therefore holds constant all context-insensitive lexical properties, ruling these factors out as an explanation for hypothetical preferences for certain grammatical systems.

\subsection{Estimating Processing Efficiency}

\subsubsection{Surprisal and Information Density}
\label{sec:surprisal}

We estimate surprisal by means of a trigram model, 
which conditions a word's probability on the 
previous two words.  For example, probability of the sentence in 
Figure~\ref{fig:dep} would be modeled as:
\[ P(\text{When}\mid \langle s\rangle)
P(\text{the} \mid \langle s\rangle \text{ When} )
P(\text{man} \mid \text{When} \text{ the} )
P(\text{arrived} \mid \text{the} \text{ man} )
\cdots
\]
where $\langle s\rangle$ indicates a sentence boundary.

N-gram models of this type are widely used in speech recognition \cite{Jel97,gold2011speech} and machine translation \cite{koehn2009}. N-gram models like the ones used here are also known to provide good approximations to computationally far more complex language models, such as probabilistic phrase structure grammars (see e.g., \cite{Genzel2002}).  One reason for this is presumably that the local context of a word often captures many semantic phenomena through the co-occurrence of related words (e.g., {\em read} and {\em book} in the trigram {\em read the book}). Trigrams also capture local syntactic patterns, such as the requirement of accusative case after certain prepositions (e.g., {\em to me})  or subject-verb agreement (e.g., {\em man arrives}).

N-gram models also have two properties that make them particularly appealing for the current purpose. First, estimating n-gram probabilities from corpora is far less computationally complex than estimating the same probabilities from structurally more complex models (such as probabilistic phrase structure grammars). Since, as we detail below, this modeling needs to be repeated many times for each language, computational simplicity is critical for the current study. Second, n-gram models have also been successfully used as models of human language processing \cite{Boston2008,Frank2011,Jaeger2010}. In fact, recent studies have argued that models that primarily rely on the information captured by local context (such as the two preceding words) fair better in explaining word-by-word variation in human processing times than structurally more complex models (\cite{Frank2011}; but see also \cite{Fossum2011, Schijndel2015}). Indeed, the finding we mentioned above, that a word's probability in context is log-linearly related to the processing difficulty it causes, was based on a trigram estimate of the type employed here \cite{Smith2013}. In short, trigram models are well-suited for the current purpose of estimating processing efficiency. One reason for this might be that human language processing preferably relies on more local information --for example, because non-local information will tend to be less informative or because non-local information will be more costly or less reliably retrieved from memory (consistent with the observation that non-local dependencies are harder to process).

In order to obtain reliable estimates of a word's trigram probability even when the preceding two words were rarely (or never) observed in the training corpus, we smooth our trigram probabilities using
the interpolated Kneser-Ney method \cite{kneser95,chen-goodman-99}. Kneser-Ney
is a technique that assigns probability to unseen n-grams
according to a measure of how likely the words in the trigram
are to combine with new words. Using Kneser-Ney smoothed trigram probabilities have two advantages over alternative n-gram models. 
First, Kneyser-Ney smoothing perform well across a wide variety of tasks and is considered one of the most effective methods of dealing with unobserved trigrams. Second, it is specifically Kneser-Ney smoothed trigram estimates of surprisal that recent work found to be linearly correlated with reaction times \cite{Smith2013}. This makes this particular approach well-suited for our purpose of estimating the average processing efficiency of a language.

Surprisal and information density can be estimated at different levels of linguistic 
description. For example, in the psycholinguistic literature on sentence processing, surprisal is usually calculated per word \cite{Frank2011, Levy2008, Smith2013}. However, psycholinguistic research on phonetic production has also calculated information density at the sub-lexical level (e.g., the information per sound in a word, \cite{CohenPriva2008, vanSon2003}). Natural languages could theoretically be efficient at one level but not the other.  

Here, we consider two estimates of information density. The first estimate is the by-word information density based on the unnormalized per-word information derived from the trigram model. This is essentially the same measure that has found to correlate linearly with word-by-word reading times in English \cite{Smith2013}. 

The second estimate is a normalized by-character estimate of the amount of information per sound or writing unit. For this second estimate, we first counted the number of unique characters in the data base (see Data above). Specifically, we used the logarithm to base 2 of that count, thereby measuring the number of bits one would need to encode all unique characters observed in the databased for each language. For example, there were 48 unique characters (5.6 bits) in our English corpora (this includes special symbols like \$) and 4394 unique characters (12.1 bits) in our Mandarin database. We then normalized the information content of each word by the number of letters in that word multiplied by the per-character bits for that language. This normalization has the advantage that it applies the same standard across different writing systems. For example, Mandarin Chinese employs a logographic writing system, so that there are no letters. For spoken language, our normalization approximates the number of phonemes in a word and its spoken duration, while also taking into account the number of distinct sounds in the language.  For written language, our normalization corresponds directly to information per character, taking into account the number of distinct symbols used in the database. 

We note that our results are not sensitive to the choice of normalization: all results were qualitatively similar without any length normalization. Furthermore, the specific normalization procedure chosen here only affects comparisons across languages (which is not of theoretical interest here), as the normalization constant does not vary within one language (see Equation \ref{eq:normalization} below, where only $|w_i|$ varies by word, whereas the per-character bits are a constant factor).

\subsubsection{Dependency Length}
\label{sec:deplength}

Our other measure of processing difficulty is dependency
length.  This metric can be read off the dependency trees,
counting the number of words from each modifier to its head
in the linear order of the sentence. For example, for the dependency structure in \figref{fig:dep}, the word {\em left} is the fifth word from the word {\em when}. The length of the SBAR$>$S dependency between {\em when} and {\em left} is thus of length 5. The SBJ$>$S dependency between the words {\em I} and {\em left}, on the other hand, is of length 1.
In our experiments, we compute the average length of all the 
dependencies in all sentence.  In \figref{fig:dep}, we have dependencies of length 1, 1, 1, 3, and 5, for an average length 
of 2.2.

A number of different metrics have been proposed to measure dependency length. For example, dependency length is sometimes measured in terms of the number of intervening non-discourse given referents \cite{Gibson2000}, or in terms of the syntactic complexity of intervening material. All of these measures tend to be {\em highly} correlated \cite{Szmrecsanyi2004, Wasow2002}. For the current purpose, we measure dependency length in words (following \cite{Ferrer2004, GildeaTemperley-cogsci10, Hawkins2004, Hawkins2007, Lohse2004, Ros2015}). This measure has the advantage that it is easy to calculate and achieves broad-coverage (see also \cite{Demberg2008}).


\subsection{Estimating Chance}
\label{sec:chance}

To obtain a chance baseline against which to compare the processing efficiency of each language, we create 1000  variants for each language. Specifically, we obtain 1000 pseudo-grammars by randomly re-ordering the dependency structures described above, while keeping the dependency relations between heads and their dependents intact. Each pseudo-grammar thus describes a theoretically possible reordering of the actual human language. Critically, this variant holds constant:

\begin{itemize}
	\item {\em all} context-insensitive lexical properties, including all semantic and phonological factors at the level of the word
	\item the number and identity of the sentences in the corpus
	\item the number of words in each sentence (which is known to affect estimates of the per-word information) and in the corpus
	\item the identity of the words in each sentence (including their part of speech) and in the corpus
	\item the number of heads, dependents, and dependencies in each sentence and in the corpus
	\item the frequency of different types of dependencies in each sentence and in the corpus
\end{itemize}

We then measure the average information density and dependency length of each variant of a language, allowing us to compare the information density and dependency length of the {\em actual} languages against what is expected by chance (i.e., against the distribution of information density and dependency lengths observed for the 1000 pseudo-grammars derived from that language). 

For our representation of a possible
fixed order, we use  {\em weighted grammars} 
\cite{GildeaTemperley-cogsci10}.  In this representation,
each dependency type
(e.g., SBJ$>$S in \figref{fig:dep}
is assigned a numeric weight $\lambda_i$ between -1 and 1. 
The head itself always has weight zero. Dependencies with negative weights appear to the 
left of the head, and dependencies with positive weights to the right. For all studies reported below, these weights were {\em held constant} for each dependency type. More specifically, we held orders constant within each {\em set of dependencies}, where a set refers to all dependency types that end in the same head. One example of a dependency set are all dependencies that end in a head noun (i.e., all noun phrase-internal dependencies). Weights thus define a deterministic order over all dependents of a head, from  left to right in order of their numeric weights. For example, with regard to the head of the sentence (S), a given pseudo-grammar might define the order SBJ SBAR {\bf S} PP NP. The relative order for the four dependency types in the rule above (SBJ$>$S, SBAR$>$S, PP$>$S, and NP$>$S), then also implies an order of SBJ {\bf S} NP for sentences in which only these two dependencies connect to S. As we show in Control Study 1, this is a conservative assumption for the calculation of chance for both information density and dependency length,  i.e., it biases {\em against} the hypothesis tested here.  

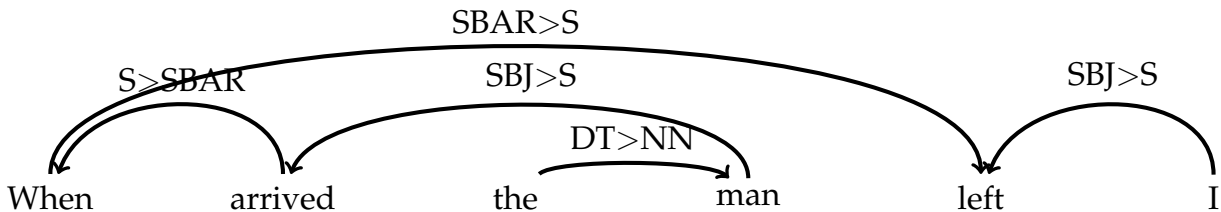
\begin{figure}
\begin{center}
\resizebox{\linewidth}{!}{
\begin{tikzpicture}[scale=1]
\draw (0,0) node(1) {When};
\draw (3,0) node(4) {arrived};
\draw (6,0) node(2) {the};
\draw (9,0) node(3) {man};
\draw (12,0) node(6) {left};
\draw (15,0) node(5) {I};
\draw[line width=1.5pt,->] (2) .. controls +(0.5,0.5) and +(-0.5,0.5) ..node[above]{DT$>$NN} (3);
\draw[line width=1.5pt,->] (3) .. controls +(0,1.5) and +(0.5,1.5) ..node[above]{SBJ$>$S} (4);
\draw[line width=1.5pt,->] (4) .. controls +(0,1.5) and +(0.5,1.5) ..node[above]{S$>$SBAR} (1);
\draw[line width=1.5pt,->] (1) .. controls +(0,2.5) and +(0,2.5) ..node[above]{SBAR$>$S} (6);
\draw[line width=1.5pt,->] (5) .. controls +(0,1.5) and +(0.5,1.5) ..node[above]{SBJ$>$S} (6);
\end{tikzpicture}}
\end{center}
\caption{Word order of a pseudo-grammar for the same sentence shown in \figref{fig:dep}.\label{fig:pseudo}}
\label{fig:graph}
\end{figure}

An example of a possible re-ordering of the 
example sentence of Figure~\ref{fig:dep} is shown in 
Figure~\ref{fig:pseudo}.

For each pseudo-grammar specified
by a set of weights $\lambda$, we estimate the
information density and dependency length
with the following procedure:

\begin{enumerate}
\item Order the training portion of our corpus
according to $\lambda$.
\item Estimate a Kneser-Ney trigram language model $L(\lambda)$ from the 
training corpus.
\item Order the test portion of our corpus
according to $\lambda$.
\item\begin{enumerate}
\item Compute the average per-word information, $h_{word}$, and normalized per-character information, $h_{character}$, in the test data according to $\lambda$, where $N$ is number of word tokens in the database, $w_i$ is the $i$th word token in the test data, $w_{i-2},w_{i-1}$ are the two preceding word tokens, and $P_{L(\lambda)}$ is the probability according to $\lambda$:
\begin{align}
 h_{i}(\lambda) &=  \log_2 \frac{1}{P_{L(\lambda)}(w_i \mid w_{i-2}, w_{i-1})}  \\
  \bar{h}_{word}(\lambda) &=  \frac{1}{N} \sum_{i=1}^{N}{h_i(\lambda)}
\end{align}
$\bar{h}_{word}$ is thus the average per-word information of a language sample, which we refer to below as the by-word information density. And for the by-character estimate of information density:
\begin{align}\label{eq:normalization}
 \bar{h}_{character}(\lambda) &= \frac{1}{N}\sum_{i=1}^N \frac{1}{|w_i|\log_2 K} h_i(\lambda)
\end{align}
where $|w_i|$ is the length of $w_i$ in characters, and $K$
is the number of unique characters in the database.
\item Compute the average dependency length of each word in the test data.
\end{enumerate}
\end{enumerate}

In all experiments, we use 9/10s of the available data as
training data in step 1 above, and the remaining 1/10 as
test data in steps 4 and 5.
This procedure takes several hours (a few minutes per random order) of computer time,
as it involves building a large table of n-gram counts
for each new random order considered.

\begin{figure}
\begin{center}
\subfigure{
	\includegraphics[width=4.5in]{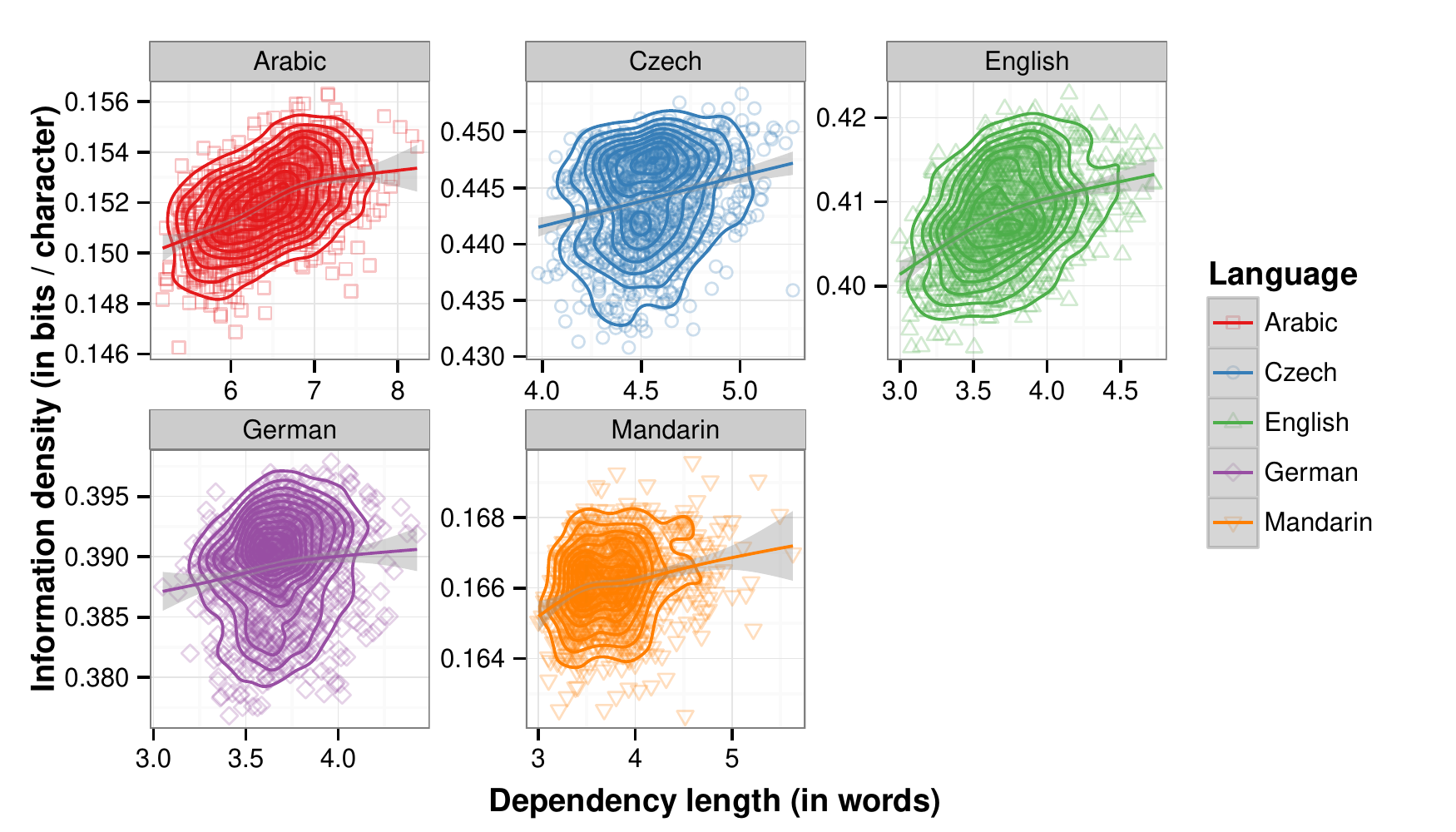} 
	\label{fig:infodep:character}
}
\subfigure{
	\includegraphics[width=4.5in]{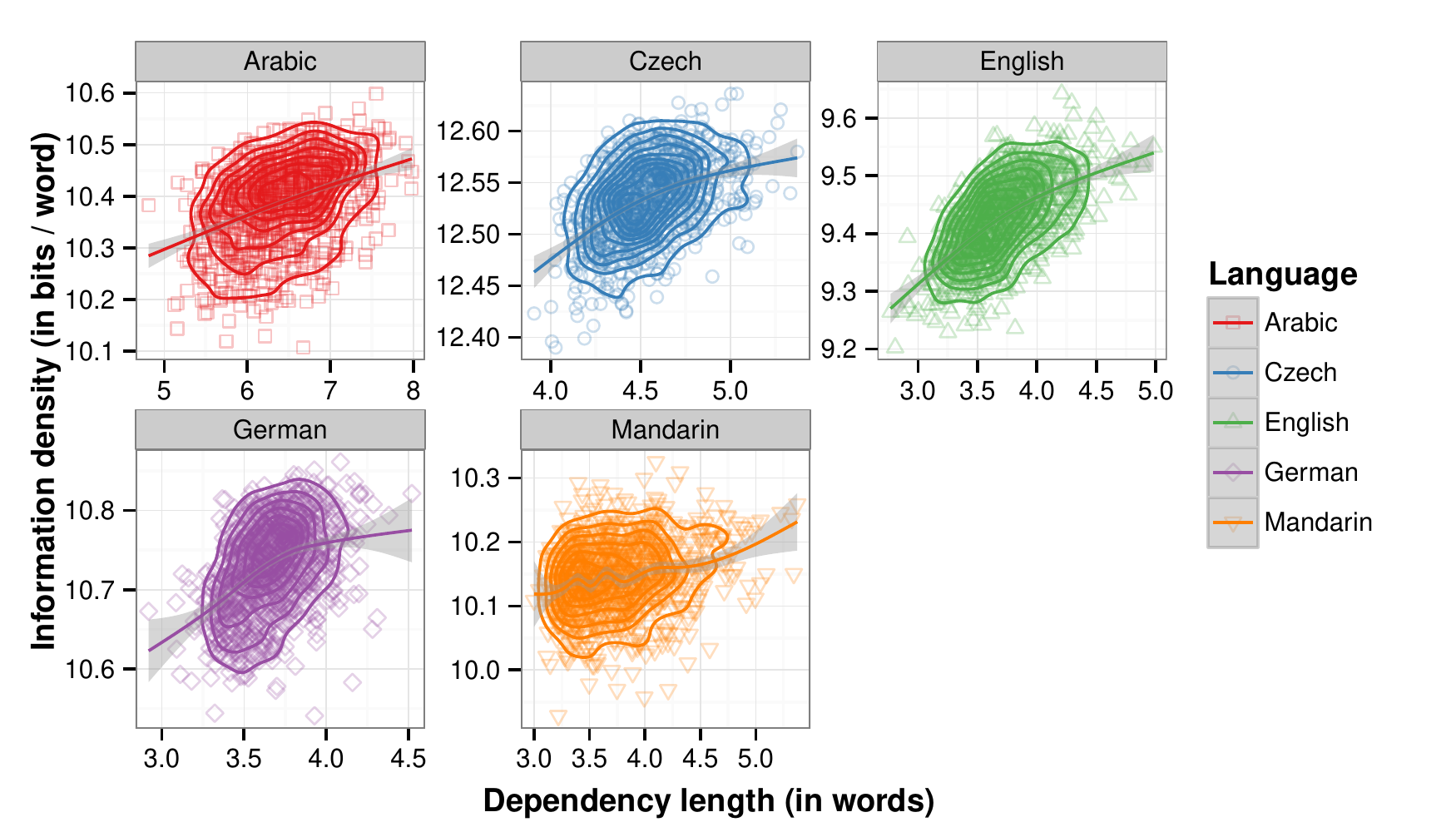} 
	\label{fig:infodep:word}
}
\caption{Average information density and dependency length of randomly generated pseudo-grammars. Individual data points show the 1000 samples for each language. Contour lines show a 2D density estimation based on a bivariate normal kernel. This summarizes the distribution of the random pseudo-grammars. Panel \subref{fig:infodep:character} shows this for the by-character estimate of information density and Panel \subref{fig:infodep:word} for the by-word estimate of information density.}\label{fig:infodep}
\end{center}
\end{figure}

\figref{fig:random} shows the information density and dependency length of all 1000 samples for the five languages. As indicated by the non-parametric smoother in \figref{fig:random}, information density and dependency length are positively correlated in the random pseudo-grammars. Although the strength of this correlation differs across languages (see Table~\ref{t:corr}), this correlation is significant in all languages (Pearson correlation $p$s $< 10^{-6}$). This means that shorter dependencies (i.e., keeping words that belong together adjacent to each other) also {\em tend} to reduce information density. This correlation makes intuitive sense. Recall that we are using a trigram language model to estimate information density. To the extent that the syntactic dependencies annotated in the corpora we employed (see Data above) capture relevant statistical dependencies between words, it is thus expected that trigram probabilities will be higher (and information density estimates lower) for word orders that keep syntactic dependencies (and thus more often within the three word window). It is, however, an interesting question for future research whether the correlation we observe here holds even when more computational more complex estimates of word probabilities are used.

\begin{table*}
\begin{tabular}{l|r|r}
       &\multicolumn{2}{c}{Pearson $\rho$}\\
       &by-character &by-word\\
    
Arabic	& 0.50 & 0.41  \\
Czech	& 0.25 & 0.47  \\
English	& 0.40 & 0.62  \\
German	& 0.15 & 0.43  \\
Mandarin& 0.23 & 0.24  \\
\end{tabular}
\caption{Correlation of information density and dependency length in the random samples created for each language.}\label{t:corr}
\end{table*}

\section{Results}
\label{sec:results}

We first compare the actual information density and dependency length of five languages in our sample against the pseudo-grammars derived from them. Then we present four control studies that serve to illustrate the robustness of our results.

\subsection{Study 1: Comparing the information density and dependency length of human languages to chance}
\begin{figure}
\begin{center}
\subfigure{
\includegraphics[width=4.5in]{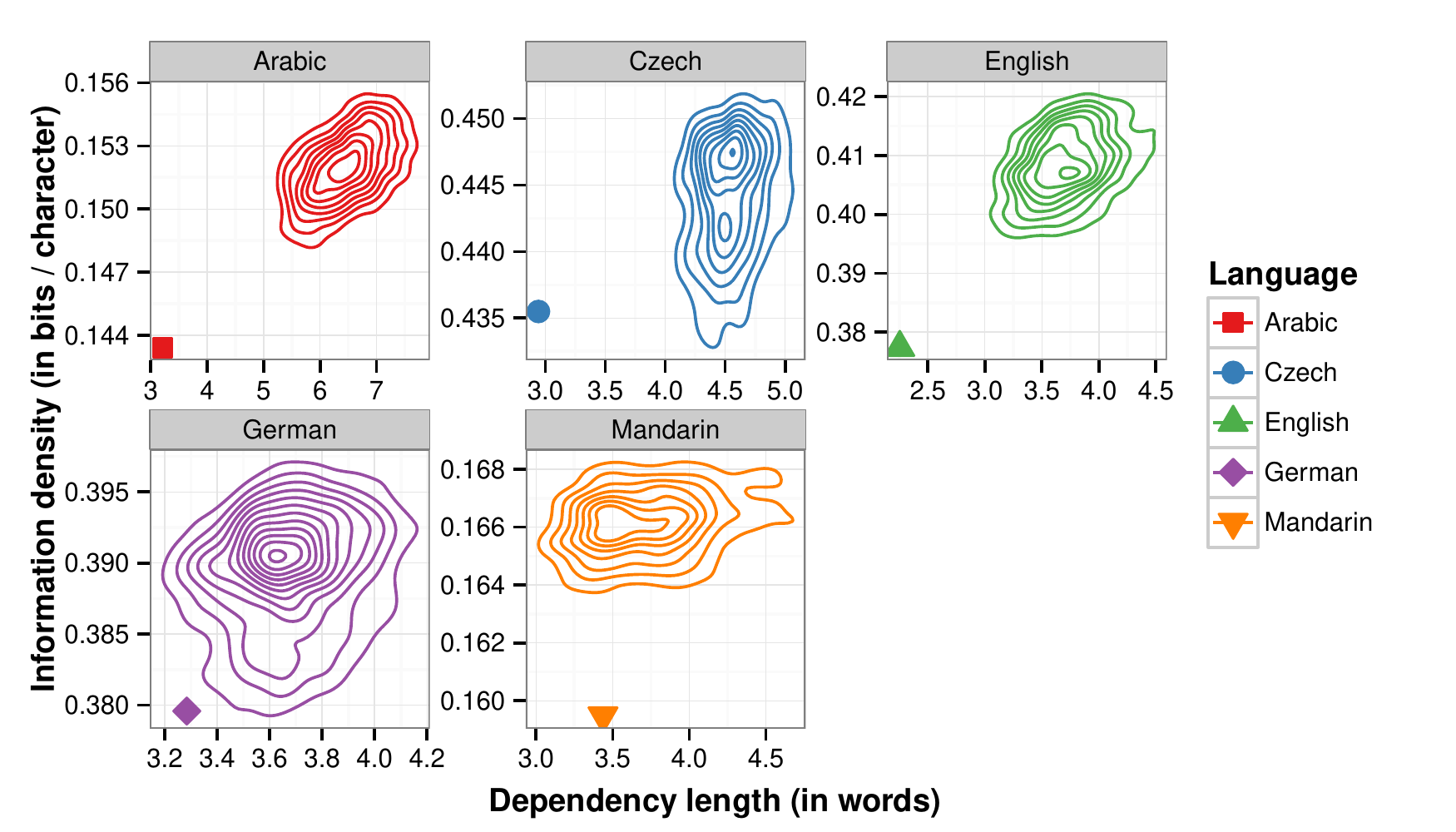}
    \label{fig:random:subfig1}
}
\subfigure{
  \includegraphics[width=4.5in]{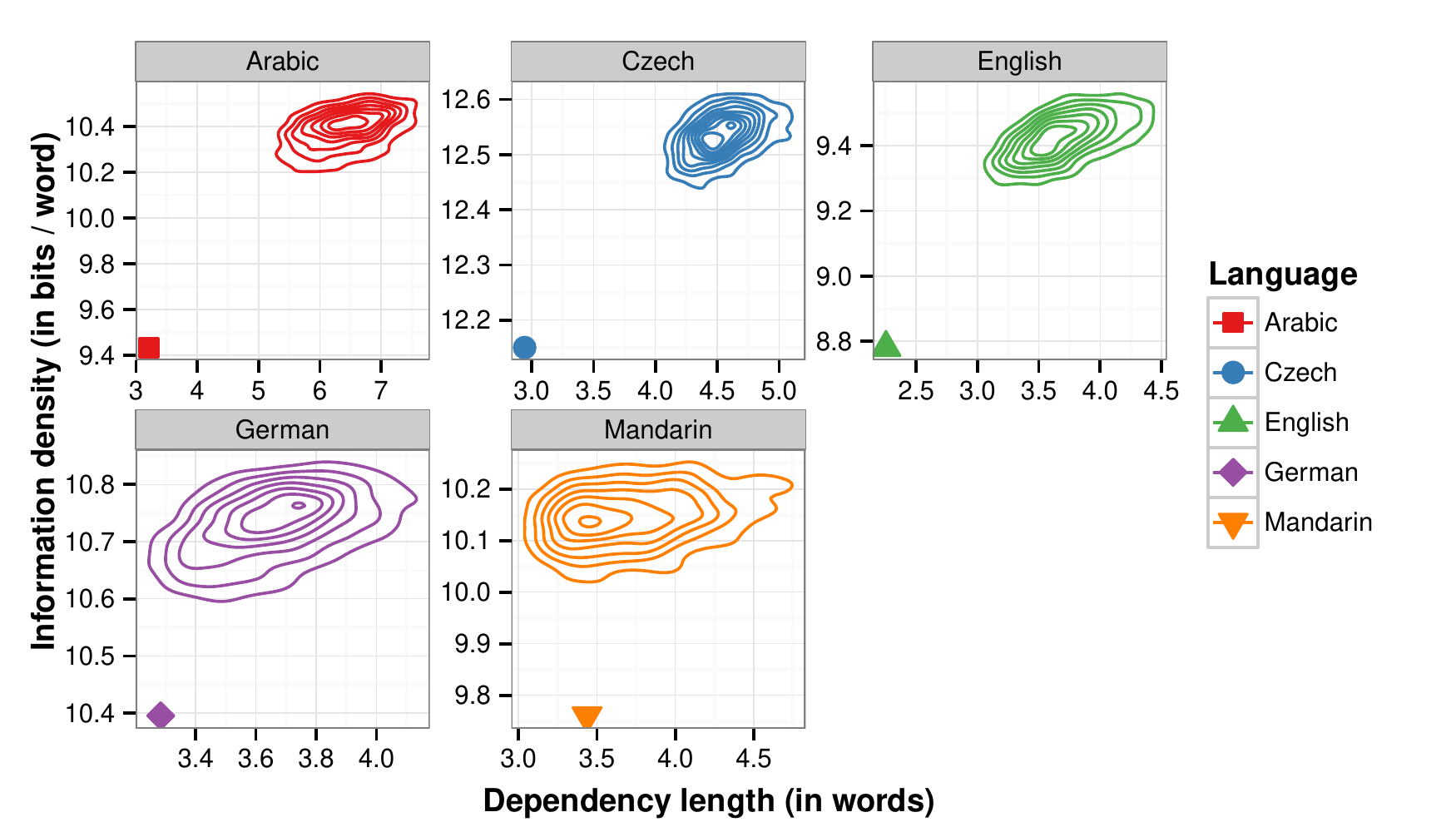}
    \label{fig:random:subfig2}
}
\caption{\label{fig:random} Illustration of the processing efficiency of actual human languages (solid shapes) compared to their processing efficiency expected by chance (contour lines). Processing efficiency is measured in terms of information density (y-axis) and dependency length (x-axis), which are both known to be positively correlated with processing times. Processing efficiency is thus higher, the lower the average information density and the lower the average dependency length. Contour lines show a 2-dimensional density estimation based on a bivariate normal kernel, summarizing the distribution of the random pseudo-grammar. Panel \subref{fig:random:subfig1} uses {\em by-character} information density. Panel \subref{fig:random:subfig2} uses {\em by-word} information density. 
}
\end{center}
\end{figure}

\figref{fig:random} shows both the actual human languages and the 1000 random samples for each of them on a plane defined by the two measures 
of processing efficiency considered here. \tableref{t:random} provides a numerical summary. As can be seen, the processing efficiency of {\em actual} Arabic, Czech, English, German, and Mandarin Chinese is considerably better than expected by chance. Specifically, applying a standard significance criterion of $\alpha = .05$, all five languages have lower information density than expected by chance, and all languages but Chinese have shorter dependency lengths than expected by chance. 

\begin{table*}
\begin{center}
\resizebox{\textwidth}{!}{
\begin{tabular}{l|rr|rr|rr|rr}
       &\multicolumn{4}{c|}{Average information density} & \multicolumn{2}{c|}{Average dependency length}\\
       &\multicolumn{2}{c|}{by-character} &\multicolumn{2}{c|}{by-word}  &  &\\
    & actual               &    higher than   & actual              & higher than & actual              & higher than\\
   & language    &   random            & language &  random & language &  random\\
Arabic	& 0.143 & 0/1000 & 9.434 & 0/1000 & 3.21 & 0/1000 \\
Czech	& 0.435 & 39/1000 & 12.150 & 0/1000 & 2.94 & 0/1000 \\
English	& 0.377 & 0/1000 & 8.781 & 0/1000 & 2.25 & 0/1000 \\
German	& 0.380 & 19/1000 & 10.395 & 0/1000 & 3.28 & 16/1000 \\
Mandarin& 0.159 & 0/1000 & 9.760 & 0/1000 & 3.44 & 220/1000 \\
\end{tabular}
}
\caption{Mean information density and dependency length of actual human language (on test data) compared to 1000 random pseudo-grammars of that language when constituent order is assumed to be fixed within each dependency type.}\label{t:random}
\end{center}
\end{table*}

Next, we present three control studies that demonstrate the robustness of our findings. Since the 
studies we present here are computationally demanding, we limit our control studies to one of the two information density estimates. We chose to focus on the per-character estimate, as we take it to be less reflective of properties specific to the writing systems of the language (such as what constitutes a written word).\footnote{For example, whereas compounds are generally written as one word in German (e.g., {\it Rotwein}), they tend to be written as separate words in English (e.g., {\it red wine}). The per-character estimate of information density is not affected by this orthographic decision.} Additionally, this is the more conservative approach given the results in \tableref{t:random}, which are stronger for by-word information density.

\subsection{Control study 1: Fixed vs. flexible constituent order}
\label{sec:headedness}

The results in \tableref{t:random} are based on pseudo-grammars that were calculated under the assumption that languages 
have fixed constituent orders within a dependency type. Interestingly, this assumption does approximate, but not quite match, what is observed for human languages. \tableref{t:actualheadedness} provides a measure of the word order consistency of the languages in our sample. 

\begin{table*}
\begin{tabular}{l|r}
       & Percentage\\
Arabic & 92.7\% \\
Czech & 74.5\% \\
English & 92.8\% \\
German & 88.7\% \\
Mandarin & 90.1\% \\
\end{tabular}
\caption{Average percentage of most frequent ordering within a dependency set, averaged across all dependency sets, for the languages in our sample.}\label{t:actualheadedness}
\end{table*}

For the calculation of chance for information density, the fixed-order assumption made in Study 1 is expected to be conservative, biasing against the hypothesis we are testing: On average, fixed constituent orders increased the predictability of words, thereby lowering the average information density. This should give the pseudo-grammars derived for Study 1 a distinct advantage compared to the actual human languages, which often do not have fixed constituent orders (or at least not {\em entirely} fixed orders). For example, even English, which is considered a relatively fixed order language, allows constituent order variation. Most obviously this holds for alternations, such as the choice between active and passive or heavy noun phrase shift (e.g.,{\it he put the book on the table} vs. {\it he put on the table the book}, but {\it he put the book he had gotten from a long lost friend on the table} vs. {\it he put on the table the book he had gotten from a long lost friend}). Generally, the assumption of fixed constituent orders in Study 1 should thus be conservative with regard to information density.

However, for the calculation of chance for dependency length, the consequences of the assumption of fixed constituent order are less clear. It is possible that this assumption made the dependency length results anti-conservative.  We therefore repeated Study 1 while allowing constituent order to vary absolutely freely. That is, rather than to use the weighted grammar approach described above in creating random pseudo-grammars, we randomly ordered all dependents for each instance of a dependency. 

\tableref{t:randomheadedness} summarizes the results. For all languages in our sample, both the by-character information density and dependency length of actual human languages were better than that observed for any of the 1000 pseudo-grammars. Control Study 1 thus replicates the results of Study 1 and shows that the assumption of fixed constituent order made in Study 1 biases against our hypothesis, relatively to allowing constituent order freedom. 

\begin{table*}
\begin{tabular}{l|rr|rr|rr}
       &\multicolumn{2}{c|}{Average information density} & \multicolumn{2}{c|}{Average dependency length}\\
       &\multicolumn{2}{c|}{by-character} & &  &\\
    & actual               &    higher than   & actual              & higher than\\
   & language    &   random             & language &  random\\
Arabic & 0.143 & 0/1000 & 3.21 & 0/1000 \\
Czech & 0.435 & 0/1000 & 2.94 & 0/1000 \\
English & 0.377 & 0/1000 & 2.25 & 0/1000 \\
German & 0.380 & 0/1000 & 3.28 & 0/1000 \\
Mandarin & 0.159 & 0/1000 & 3.44 & 0/1000 \\
\end{tabular}
\caption{Mean information density and dependency length of actual human language (on test data) compared to 1000 random pseudo-grammars of that language when constituent order is completely randomized.}\label{t:randomheadedness}
\end{table*}

We further note that the results of Study 1 were also replicated when constituents were allowed to order freely, but the position of dependents relative to the head was held constant (e.g., if all dependents occurred to the right of their head). Taken together, this suggests that the results obtained in Study 1 are robust to assumptions about constituent order freedom in the calculation of the chance baseline.

\subsection{Control study 2: Sensitivity to Genre and Mode}

While our primary datasets are taken from newspaper
text, we wanted to test whether our results were sensitive 
to the genre of the corpus, and in particular whether
edited, written text might have different properties than
spontaneous, spoken text. Unfortunately, large syntactically annotated corpora are available only for very few languages. 
Here we test our hypothesis against conversational speech data from English.

Repeating Study 1 on the English Switchboard corpus of conversational speech, we again find that the processing efficiency of actual English  is better than expected by chance. Actual conversational English had better per-character information density than all of 1000 random word orders, and better dependency length than all of 1000 random orders (both $p$s $< .0001$).

\subsection{Control study 3: Sensitivity to Corpus Size}
\label{sec:small}

Finally, we tested the sensitivity of our results to the amount of text available for estimating the parameters of
the Kneser-Ney trigram model.  We thus repeated the analysis reported above, using a much smaller
data set of 1000 sentences randomly drawn from the Wall Street Journal (i.e., about 2.5\% of the original corpus). Unsurprisingly, information density estimates were higher compared to the main study (reported in \tableref{t:random})--this is a direct consequence of the reduced data size: for smaller corpora, there will be more 
n-grams in the test data that were never observed in the training data; these n-grams are assigned low probability (and thus high information). The estimates based on a smaller corpus are also expected to be less reliable because a large porportion of 
the n-grams in the test data will be unseen in the training
data, regardless of the word order used.  To quantify this effect,
using actual English word order, we find that with 1000 training
sentences, only 10\% of bigram tokens in test data have been 
observed in training data, even when not predicting any
single words in the test data that are unseen in training data.
In contrast, with our full training set for English, the corresponding 
figure is 31\%.
Despite the low coverage of n-grams in our 1000-sentence 
training set, we again find that 
actual English has better per-character information density
than all of 1000 random word orders, and better dependency
length than all of 1000 random orders (both $p$s $< .05$). 

\subsection{Summary}
We find that the five languages we investigated all have significantly higher processing efficiency than expected by chance. This holds for both measures of processing efficiency considered here. For information density, all five languages fall into the top 95th percentile or better. For dependency length, four of the five languages fall into the 95th percentile, and one language (Mandarin) falls into approximately the 75th percentile of the distribution defined by the random pseudo-grammars. Our findings held regardless of the size of the corpus and, more importantly, for both written and spoken language. Taken together, this suggests that language use --specifically, pressures that originate in the incremental processing of language-- shape the grammar of languages over time. 

We note that information density and dependency length were not independent in our random samples. It is thus possible that what we have --following the literature-- treated as two independent measures of processing efficiency is in reality due to one underlying cause. This would not affect the conclusion that the processing efficiency of human languages is better than expected by chance. Further, it is worth noting that the correlations in \tableref{t:corr} are mild. Indeed, Study 2 finds that information density and dependency length can be optimized separately.

\section{Study 2: Are natural languages optimal with regard to information density and dependency length?}

Next, we tested whether an even stronger claim can be made. Specifically, we wondered whether the pressures for efficient processing are sufficiently strong to constrain language change 
to the subspace of possible grammars that is optimal (or very close to optimal) in terms of processing efficiency. As outlined in the introduction, many pressures of language use have been hypothesized to bias and constrain 
language change, thereby contributing to cross-linguistically observed properties of languages. We thus did not expect languages to be optimal in terms of processing efficiency. We begin by describing the procedure used to estimate the minimum possible information density for each language. Then we describe the procedure used to estimate the minimum possible dependency length for each language. Finally, we present a procedure that jointly optimizes both information density and dependency length for each language, allowing us to compare human languages against pseudo-grammars that optimally trade-off the two major contributors to processing efficiency. The results of these three calculations are presented and discussed at the end of this section.

\subsection{Computing pseudo-grammars with optimal information density}

We begin by describing the procedure used to calculate the minimum possible information density $h^*$ for each language:
\[ h^* = \min_\lambda h(\lambda) \]

In order to find $h^*$, we optimize one weight at a time,
holding all others fixed, and iterating though the set
of weights to be set.  The objective function describing 
information density is piecewise constant, 
as the objective function will not change until one weight 
crosses some other, causing two dependents to reverse order,
at which point the objective will discontinuously jump.
This non-differentiability implies that methods based on gradient
ascent will not apply.  However, because the objective function
only changes at points where one weight crosses another's value,
the set of segments of weight values with different values 
of the objective function can be exhaustively enumerated.
In fact, the only significant points are the values of other
weights for dependency types which occur in the corpus attached
the same head as the dependency being optimized.  We 
build a table of interacting dependencies as a preprocessing
step on the data, and then when optimizing a weight, 
consider the sequence of values between consecutive 
interacting weights.  
For each value in this sequence, we evaluate the objective
function $h$ on the test corpus, and choose the value 
yielding the minimum value of $h$.

This search procedure is similar to one previously used
for finding the grammar that minimizes dependency length \cite{GildeaTemperley-cogsci10}.
Our present problem, however, is considerably more computationally intensive,
because when evaluating each possible value of each weight,
we must examine the {\em entire} test corpus, whereas,
when optimizing dependency length, one can take a shortcut in
evaluation, by only considering sentences with the dependency
type whose weight has been modified.
In our case, since all the parameters of the n-gram language
model are subject to change at each step, we must evaluate the
language model on every word in the test corpus.

This optimization process is not guaranteed to find the global
maximum, but is guaranteed to converge simply from the fact
that there are a finite number of objective function values, 
and the objective function must increase at each step at which weights
are adjusted.  Running the optimization procedure from different
random initializations, we find that, while final grammars
are not identical, they are very close in terms of our objective
function, which indicates that we are likely to be close to 
the global optimum.  For example, in ten runs optimizing 
the by-word information density of English, our final values of the 
objective function have a variance of less than $10^{-5}$.

For our experiments, we find that
the optimization procedure converges after several days of computer 
time.  However, the procedure reaches points very close 
to the eventual optimum within several hours.

\subsection{Computing pseudo-grammars with optimal dependency length}
We also optimize our pseudo-grammar in order to 
find the weights giving the lowest dependency length:
\[ d^* = \min_\lambda d(\lambda) \]
where $d(\lambda)$ is the average dependency length
for the pseudo-grammar with weights $\lambda$.
The search over weights uses the same algorithm described above.

\begin{figure}
\begin{center}
  \includegraphics[width=6.5in]{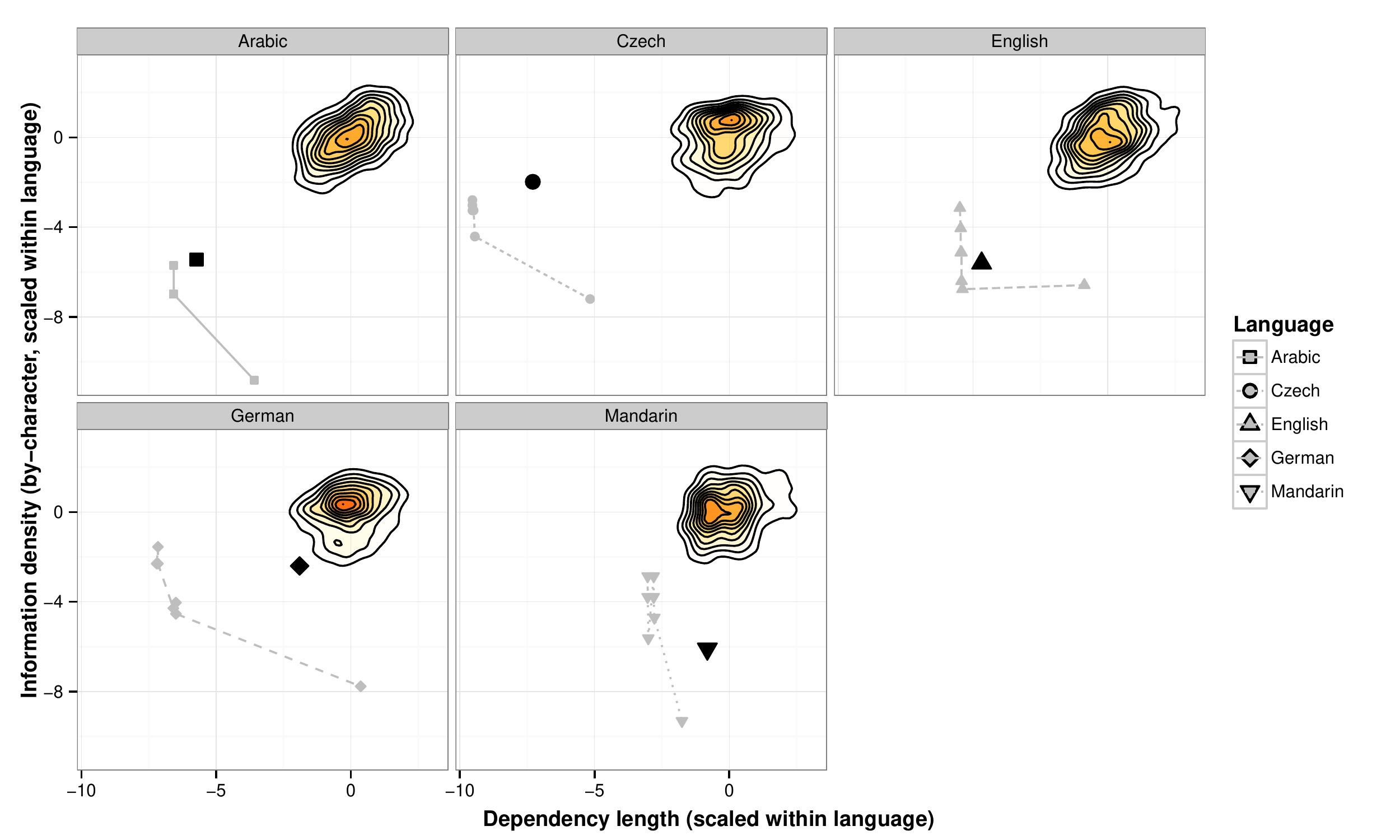}
\caption{\label{fig:optimized} Comparison of human languages to grammars that {\em optimize} information density and dependency length. Gray-shaded lines show grammars that are optimal under a given trade-off between information density and dependency length (bottom-most point: grammar with optimal information density; left-most point: grammar with optimal dependency length; in between: grammars that optimized weighted sums of information density and dependency length). To facilitate cross-language comparison, information density is normalized by character and both information density and dependency lengths are standardized, so that axes represent $z$-values. Contour lines and fill show the distribution of the random reorderings of {\em all} languages combined. 
}
\end{center}
\end{figure}

\subsection{Joint Optimization}

There may be a trade-off between information density and dependency length. Although we found information density and dependency length to be positively correlated in the random pseudo-grammars generated for Study 1, these correlations were mild to moderate. It is therefore possible that optimization of information density trades off against optimization of dependency length (and vice versa). We therefore investigate the effect on the dependency length of optimizing for information density, and vice versa. We also experiment with a joint objective function $j$ that
combines information density and dependency length:
\[ j^* = \min_\lambda (1-\alpha)d(\lambda) + \alpha h(\lambda) \]

We then applied the same optimization algorithm described above to this joint measure of processing efficiency. The separate optimizations described in the previous two sections correspond to $\alpha$s of 1 and 0, respectively. We also considered $\alpha$s of $.5$, $.6$, $.7$, $.8$, and $.9$ (note that these weights are hard to interpret by themselves: information density and dependency length are on different scales, as shown in \tableref{t:random} above.

\subsection{Results}

Table~\ref{t:joint} shows results when optimizing for 
different objective functions in each column. Rows show the by-character information density and dependency length for each language. The left half of the table shows the information density and dependency length for the optimal pseudo-grammars derived by optimizing information density (ID), dependency length (DL) or both jointly with $\alpha = 0.5$ (ID \& DL). The right half of the table shows the information density and dependency length of the actual human language, as well as the mean of the random pseudo-grammars generated for Study 1.

Our first observation from \tableref{t:joint} is that optimizing either information density or dependency length indeed comes at the expense of the other property (despite the overall {\em positive} correlations between information density and dependency length in the random pseudo-grammars, cf.\ \figref{fig:infodep}). Further, the average information density and dependency length of all five natural languages is overall closer to the joint optimum, than to either of the separate optima, suggesting that natural languages indeed trade-off information density and dependency length.

\begin{table*}
\begin{tabular}{lrrr|rrr}
          &\multicolumn{3}{c|}{Optimize for}&\\
          &{\em ID} & {\em ID \& DL} & {\em DL}  & Actual & Mean of random        \\
          &    &                 &       &language & pseudo-grammars \\
\hline
Arabic&&&&&\\
information density & 0.135 & 0.141 & 0.143 & 0.143 & 0.146 \\
dependency length   & 4.42 & 2.73 & 2.73 & 3.21 & 4.75 \\
\hline
Czech&&&&&\\
information density & 0.413 & 0.431 & 0.432 & 0.435 & 0.436 \\
dependency length   & 3.41 & 2.45 & 2.45 & 2.94 & 3.27 \\
\hline
English&&&&&\\
information density & 0.372 & 0.386 & 0.391 & 0.377 & 0.394 \\
dependency length   & 3.45 & 2.01 & 2.00 & 2.25 & 2.79 \\
\hline
German&&&&&\\
information density & 0.358 & 0.380 & 0.383 & 0.380 & 0.386 \\
dependency length   & 3.76 & 2.18 & 2.18 & 3.28 & 3.91 \\
\hline
Mandarin&&&&&\\
information density & 0.156 & 0.163 & 0.162 & 0.159 & 0.164 \\
dependency length   & 3.06 & 2.55 & 2.55 & 3.44 & 4.56 \\
\end{tabular}
\caption{Results of separate and joint optimization of information density ({\em ID}) and dependency length ({\em DL}). For comparison, the two rightmost columns provide the information density and dependency length of human languages as well as the mean of the random pseudo-grammars (repeated from \tableref{t:random}).}\label{t:joint}
\end{table*}

\begin{figure}
\begin{center}
\includegraphics[width=3.3in]{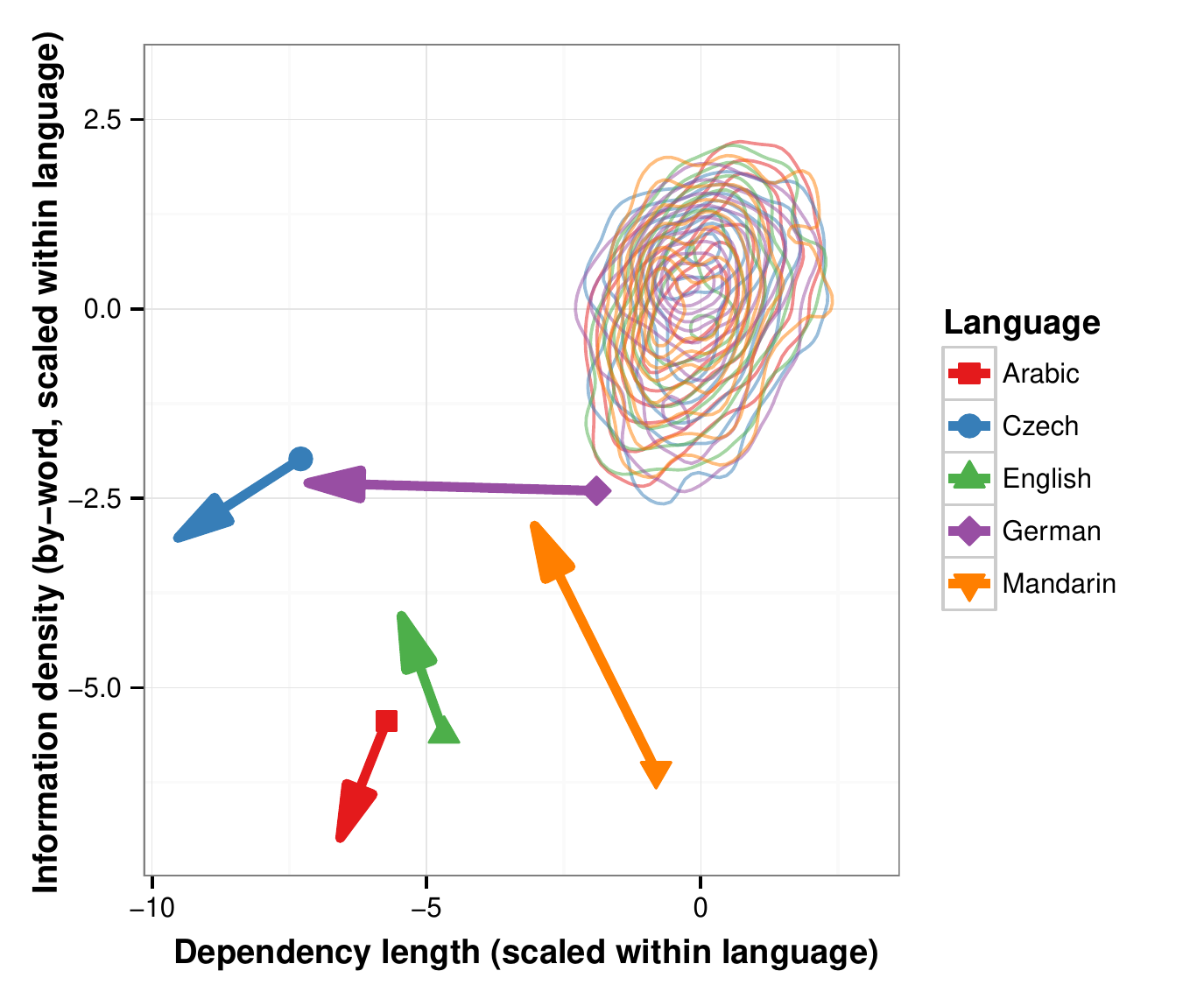} 
\caption{Comparison of human languages and optimized languages derived from them, in terms of by-character information density and 
dependency length. To put all languages on a comparable scale, the two axes are standardized (i.e., they represent $z$-values). Solid shapes show the human languages. Arrows point to optimal pseudo-grammars. Contour lines show a 2D density estimation based on a bivariate normal kernel. This summarizes the distribution of the random pseudo-grammar.}\label{fig:optimal}
\end{center}
\end{figure}

This means that the jointly optimized pseudo-grammars provide the most relevant point of comparison for natural languages, since we are interested in understanding whether the grammars of natural languages are close to optimal in their {\em overall} processing efficiency.  One way to further illustrate this trade-off is to look at the equi-weighted joint optimization ($\alpha = .5$). These jointly optimized grammars do {\em not} unambiguously outperform actual natural languages. For two of the five languages, Arabic, and Czech, the equi-weighted jointly optimized pseudo-grammar has lower information density and dependency length. For the other three languages, however, the optimized pseudo-grammar is better on one dimension of processing efficiency, but worse on the other (e.g., for the jointly optimized pseudo-grammar of Mandarin, a slight improvement in dependency length results in a considerably worsening in information density, compared to actual Mandarin). This is visualized in \figref{fig:optimal}.



The full results of the different joint optimizations,
varying the weighting parameter $\alpha$, are shown in \figref{fig:optimized}. 
If language processing is just one among many equally important factors that shape word order preferences over time, the processing efficiency of optimized grammars should far outperform that of the actual human languages.
The gray lines in the Figure \ref{fig:optimized} can be thought of as representing a `frontier' of optimality within the space of possible grammars for each of the languages in our sample.

None of the languages in our sample lies on this frontier. That is, all languages could theoretically change to have better information density and dependency length. However, we also find that the word orders of some languages (Arabic and English) have close to optimal processing efficiency. Of the five languages investigated here, only the word order of German clearly has non-optimal processing efficiency. One possible reason for such striking differences between languages might be differences in how they use other means than word order to convey the relations between words in a sentence (e.g., word-internal structure and morphosyntactic means). It is also possible that the historical development of some languages has been more strongly affected by other factors of language use (such ease of production or learnability). The current computational simulations cannot distinguish between these possibilities. The approach applied here does, however, point a way forward: as better quantitative models of these other factors become available, future simulations can investigate to what extent other aspects of language use shape the cultural evolution of language. 

We further note that our optimization procedure held constant the headedness of each dependency type. As mentioned above, this is close to, but not identical to, what the human languages in our sample do. It is an open question how a relaxation of the the constant-headedness constraint would affect our results. Varying headedness arguably makes a language harder to learn, so that the assumption of constant headedness can be seen as holding constant what likely constitutes an additional (third) constraint that languages need to balance.

\section{Discussion}
The present results suggest that language processing affects language change: all natural languages for which we could test the hypothesis have word orders that make them easier to process than expected by chance. Specifically, the average information density and dependency length of the natural languages in our sample is lower than would be expected if language change was not subject to a bias toward systems with high processing efficiency. 

To the best of our knowledge, this is the first cross-linguistic broad-coverage study of the processing efficiency of natural languages. The measures of processing efficiency we have employed here are two of the best documented correlates of processing complexity. By {\em calculating} the average information density and dependency length of natural languages based on large collections of text from these languages, we were able to side step the insurmountable challenges that would be associated with a behavioral approach to this question (see Introduction). There are three directly related previous findings that we are aware of \cite{Ferrer2004, GildeaTemperley-cogsci10, futrell2015large}. Gildea and Temperley \cite{GildeaTemperley-cogsci10} investigated the average dependency length of two closely related languages, English and German. Ferrer i Cancho \cite{Ferrer2004} investigates Czech and Romanian, while 
Futrell et al.\ \cite{futrell2015large} study 37 languages spoken worldwide.
These studies found that the languages studied had shorter average dependency length than expected by chance. 
Our contribution is to --for the first time, to our knowledge-- assess the joint effect of two of the biggest contributors to the grammatical processing efficiency of a language, information density and dependency length. As the trade-off between these factors in Study 2 shows, it is crucial to investigate the effect of multiple contributors to processing efficiency simultaneously. 

Some other recent studies complement the approach taken here. These studies have tested whether learners of miniature languages designed by experimenters prefer languages that increase processing efficiency \cite{Fedzechkina2011, Fedzechkina2012, Fedzechkina2013}. In the most informative of these studies, great care is taken to rule out native language biases as the source of the observed preferences (cf. \cite{Goldberg2013}). For example, \cite{Fedzechkina2014} finds that language learners prefer languages that reduce unnecessary uncertainty about the syntactic structure of sentences. Studies like this provide evidence that processing preferences can bias the outcome of language {\em learning} and thus provide support for one causal pathway through which processing preferences could come to affect language change, thereby shaping languages over time. 

There are several caveats that apply to our study. The most obvious perhaps is that we have only considered syntactic dependencies that are annotated in the available syntactic corpora. These dependencies constitute an impoverished subset of all the semantic and syntactic dependencies that human comprehenders process when listening or reading. For example, one obvious omission in our approach is that we did not consider the internal structure of words, the complexity of which varies starkly across languages. Another simplifying assumption we have implicitly made in our studies is the focus on information density and dependency length. While these two measure of processing efficiency are arguably the best documented ones, there are other 
properties of grammatical systems that are known to affect processing efficiency (e.g., interference in memory due to similar words or referents, \cite{Lewis2006, MacDonald2013}). As far as we can tell, neither of these simplifying assumption is likely to have biased the results in favor of the hypothesis tested here (for that to be the case, the measures of processing efficiency we have employed here would have to be systematically inversely correlated with other measures or properties of the languages under study).

\paragraph{Acknowledgments} The authors thank Masha Fedzechkina, Chigusa Kurumada, and Olga Nikolayeva for feedback on earlier versions of this paper. This work was partially funded by National Science Foundation award IIS-1446996 to DG and National Science Foundation CAREER award IIS-1150028 to TFJ. The views expressed here at not necessarily those of the funding agencies.\fj{, and audiences at the University of Geneva, the workshop on {\em Causality in the Language Sciences} held at the Max Planck Institute for Evolutionary Anthropology in Leipzig, and the workshop on {\em Information density and Linguistic Encoding} held at the Sonderforschungsbereich 1102 in Saarbr\"ucken}

\bibliographystyle{plain}
\bibliography{all}

\begin{thebibliography}{10}

\bibitem{Arnold2004}
Jennifer~E Arnold, Thomas Wasow, Ash Asudeh, and Peter Alrenga.
\newblock {Avoiding attachment ambiguities : The role of constituent ordering}.
\newblock {\em Journal of Memory and Language}, 51:55--70, 2004.

\bibitem{Arnold2000}
Jennifer~E Arnold, Thomas Wasow, Anthony Losongco, and Ryan Ginstrom.
\newblock {Heaviness vs. newness: the effects of structural complexity and
  discourse status on constituent ordering}.
\newblock {\em Language}, 76(1):28--55, 2000.

\bibitem{Baayen2006}
R~H Baayen.
\newblock {Morphological influences on the recognition of monosyllabic
  monomorphemic words}.
\newblock {\em Journal of Memory and Language}, 55:290--313, 2006.

\bibitem{Balota2004}
David~A Balota, Michael~J Cortese, Susan~D Sergent-Marshall, Daniel~H Spieler,
  and Melvin~J Yap.
\newblock {Visual Word Recognition of Single-Syllable Words}.
\newblock {\em Journal of experimental psychology: General}, 133(2):283--316,
  2004.

\bibitem{Bates1982}
Elizabeth Bates and Brian MacWhinney.
\newblock {Functionalist approaches to grammar}.
\newblock In {\em Language acquisition: the state of the art}, pages 173--218.
  1982.

\bibitem{Bates1987}
Elizabeth Bates and Brian MacWhinney.
\newblock {Competition, Variation, and Language Learning}.
\newblock In Brian MacWhinney, editor, {\em Mechanisms of Language
  Acquisition}, chapter~6, pages 157--194. 1987.

\bibitem{PDT}
A.~B{\"o}hmov{\'a}, J.~Haji\v{c}, E.~Haji\v{c}ov{\'a}, and B.~Hladk{\'a}.
\newblock The {PDT}: a 3-level annotation scenario.
\newblock In A.~Abeill{\'e}, editor, {\em Treebanks: Building and Using Parsed
  Corpora}, volume~20 of {\em Text, Speech and Language Technology}, chapter~7.
  Kluwer Academic Publishers, Dordrecht, 2003.

\bibitem{Boston2008}
Marisa~Ferrara Boston, John Hale, Reinhold Kliegl, Umesh Patil, and Shravan
  Vasishth.
\newblock {Parsing costs as predictors of reading difficulty: An evaluation
  using the Potsdam Sentence Corpus}.
\newblock {\em Journal of Eye Movement Research}, 2(1):1--12, 2008.

\bibitem{TIGER}
S.~Brants, S.~Dipper, S.~Hansen, W.~Lezius, and G.~Smith.
\newblock The {TIGER} treebank.
\newblock In {\em Proc. of the 1st Workshop on Treebanks and Linguistic
  Theories ({TLT})}, 2002.

\bibitem{buchholz2006conll}
Sabine Buchholz and Erwin Marsi.
\newblock {CoNLL-X} shared task on multilingual dependency parsing.
\newblock In {\em Proceedings of the Tenth Conference on Computational Natural
  Language Learning}, pages 149--164. Association for Computational
  Linguistics, 2006.

\bibitem{Bybee2001}
Joan Bybee and Paul~J Hopper.
\newblock {\em {Frequency and the Emergence of Linguistic Structure}}.
\newblock 2001.

\bibitem{Bybee1999}
Joan Bybee and Joanne Scheibman.
\newblock {The effect of usage on degrees of constituency: the reduction of
  don\'t in English}.
\newblock {\em Linguistics}, 37(4):575--596, 1999.

\bibitem{chen-goodman-99}
Stanley~F Chen and Joshua Goodman.
\newblock An empirical study of smoothing techniques for language modeling.
\newblock {\em Computer Speech \& Language}, 13(4):359--393, 1999.

\bibitem{Choi2007}
Hye-won Choi.
\newblock {Length and Order: A Corpus Study of Korean Dative-Accusative
  Construction}.
\newblock {\em 담화와 인지}, 14(3):207--227, 2007.

\bibitem{CohenPriva2008}
Uriel {Cohen Priva}.
\newblock {Using Information Content to Predict Phone Deletion}.
\newblock In Natasha Abner and Jason Bishop, editors, {\em Proceedings of the
  27th West Coast Conference on Formal Linguistics}, pages 90--98, 2008.

\bibitem{Collins99}
Michael~John Collins.
\newblock {\em Head-driven Statistical Models for Natural Language Parsing}.
\newblock PhD thesis, University of Pennsylvania, Philadelphia, 1999.

\bibitem{DelPradoMartin2004a}
Ferm\'{\i}n~Moscoso {del Prado Mart\'{\i}n}, Aleksandar Kosti\'{c}, and
  R~Harald Baayen.
\newblock {Putting the bits together: an information theoretical perspective on
  morphological processing.}
\newblock {\em Cognition}, 94(1):1--18, November 2004.

\bibitem{Demberg2008}
Vera Demberg and Frank Keller.
\newblock {Data from eye-tracking corpora as evidence for theories of syntactic
  processing complexity.}
\newblock {\em Cognition}, 109(2):193--210, November 2008.

\bibitem{Fedzechkina2014}
Maryia Fedzechkina.
\newblock {\em Communicative Efficiency, Language Learning, and Language
  Universals}.
\newblock PhD thesis, University of Rochester, 2014.

\bibitem{Fedzechkina2011}
Maryia Fedzechkina, T~Florian Jaeger, and Elissa~L Newport.
\newblock {Functional Biases in Language Learning: Evidence from Word Order and
  Case-Marking Interaction}.
\newblock In {\em 33rd Annual Meeting of the Cognitive Science Society}, number
  2004, pages 318--323, 2011.

\bibitem{Fedzechkina2012}
Maryia Fedzechkina, T.~Florian Jaeger, and Elissa~L. Newport.
\newblock {Language learners restructure their input to facilitate efficient
  communication}.
\newblock {\em Proceedings of the National Academy of Sciences of the United
  States of America}, pages 1--6, October 2012.

\bibitem{Fedzechkina2013}
Maryia Fedzechkina, T~Florian Jaeger, and Elissa~L Newport.
\newblock {Communicative biases shape structures of newly acquired languages}.
\newblock In M.~Knauff, N.~Pauen, N.Sebanz, and I.~Wachsmuth, editors, {\em
  Proceedings of the 35th Annual Meeting of the Cognitive Science Society
  (CogSci13)}, pages 430--435. Cognitive Science Society, Austin, TX, 2013.

\bibitem{Ferrer2004}
Ramon {Ferrer i Cancho}.
\newblock {Euclidean distance between syntactically linked words}.
\newblock {\em Physical Review E}, 70(056135):1--5, 2004.

\bibitem{Fossum2011}
Victoria Fossum and Roger Levy.
\newblock Sequential vs. hierarchical syntactic models of human incremental
  sentence processing.
\newblock In {\em Proceedings of the 3rd Workshop on Cognitive Modeling and
  Computational Linguistics}, pages 61--69. Association for Computational
  Linguistics, 2012.

\bibitem{Frank2011}
Stefan~L Frank and Rens Bod.
\newblock Insensitivity of the human sentence-processing system to hierarchical
  structure.
\newblock {\em Psychological Science}, 22(6):829--834, 2011.

\bibitem{futrell2015large}
Richard Futrell, Kyle Mahowald, and Edward Gibson.
\newblock Large-scale evidence of dependency length minimization in 37
  languages.
\newblock {\em Proceedings of the National Academy of Sciences},
  112(33):10336--10341, 2015.

\bibitem{Genzel2002}
Dmitriy Genzel and Eugene Charniak.
\newblock {Entropy Rate Constancy in Text}.
\newblock In {\em Proceedings of the 40th Annual Meeting of the Association for
  Computational Linguistics (ACL)}, pages 199--206, 2002.

\bibitem{Gibson1998}
E~Gibson.
\newblock {Linguistic complexity: locality of syntactic dependencies.}
\newblock {\em Cognition}, 68(1):1--76, August 1998.

\bibitem{Gibson2000}
Edward Gibson.
\newblock {The Dependency Locality Theory: A Distance-Based Theory of
  Linguistic Complexity}.
\newblock In Alec Marantz, Yasushi Miyashita, and Wayne O'Neil, editors, {\em
  Image, language, brain: Papers from the first mind articulation symposium},
  chapter~5, pages 95--126. 2000.

\bibitem{Gibson2013}
Edward Gibson, Steven~T Piantadosi, Kimberly Brink, Leon Bergen, Eunice Lim,
  and Rebecca Saxe.
\newblock {A noisy-channel account of crosslinguistic word-order variation.}
\newblock {\em Psychological Science}, 24(7):1079--88, July 2013.

\bibitem{GildeaTemperley-cogsci10}
Daniel Gildea and David Temperley.
\newblock Do grammars minimize dependency length?
\newblock {\em Cognitive Science}, 34(2):286--310, 2010.

\bibitem{Godfrey92}
J.~Godfrey, E.~Holliman, and J.~McDaniel.
\newblock {SWITCHBOARD}: {T}elephone speech corpus for research and
  development.
\newblock In {\em IEEE ICASSP-92}, pages 517--520, San Francisco, 1992. IEEE.

\bibitem{gold2011speech}
Ben Gold, Nelson Morgan, and Dan Ellis.
\newblock {\em Speech and audio signal processing: processing and perception of
  speech and music}.
\newblock John Wiley \& Sons, 2011.

\bibitem{Goldberg2013}
Adele~E Goldberg.
\newblock Substantive learning bias or an effect of familiarity? comment on.
\newblock {\em Cognition}, 127(3):420--426, 2013.

\bibitem{Graff2009}
Peter Graff and T~Florian Jaeger.
\newblock {Locality and Feature Specificity in OCP Effects: Evidence from
  Aymara, Dutch, and Javanese}.
\newblock In {\em Proceedings of the Main Session of the 45th Meeting of the
  Chicago Linguistic Society}, pages 1--15, 2009.

\bibitem{Grodner2005}
Daniel Grodner and Edward Gibson.
\newblock {Consequences of the serial nature of linguistic input for sentenial
  complexity.}
\newblock {\em Cognitive science}, 29(2):261--90, March 2005.

\bibitem{Guy1996}
Gegory~R Guy.
\newblock {Form and function in linguistic variation}.
\newblock In Gegory~R Guy, Crawford Feagin, Deborah Schiffrin, and John Baugh,
  editors, {\em Towards a social science of language: Papers in honor of
  William Labov. Volume 1: Variation and change in language and society}, pages
  221--252. Benjamins Publishing Compagny, Amsterdam, 1996.

\bibitem{hajic2004prague}
Jan Haji{\v{c}}, Otakar Smr{\v{z}}, Petr Zem{\'a}nek, Jan {\v{S}}naidauf, and
  Emanuel Be{\v{s}}ka.
\newblock Prague arabic dependency treebank: Development in data and tools.
\newblock In {\em Proc. of the NEMLAR Intern. Conf. on Arabic Language
  Resources and Tools}, pages 110--117, 2004.

\bibitem{Hale2001}
John Hale.
\newblock {A Probabilistic Earley Parser as a Psycholinguistic Model}.
\newblock In {\em NAACL '01 Proceedings of the second meeting of the North
  American Chapter of the Association for Computational Linguistics on Language
  technologies}, pages 1--8, 2001.

\bibitem{Hauser2002}
Marc~D Hauser, Noam Chomsky, and W~Tecumseh Fitch.
\newblock The faculty of language: what is it, who has it, and how did it
  evolve?
\newblock {\em science}, 298(5598):1569--1579, 2002.

\bibitem{Hawkins2014}
J.~A. Hawkins.
\newblock {\em Cross-linguistic variation and efficiency}.
\newblock Oxford University Press, Oxford, UK, 2014.

\bibitem{hawkins94}
John Hawkins.
\newblock {\em A Performance Theory of Order and Constituency}.
\newblock Cambridge University Press, Cambridge, UK, 1994.

\bibitem{Hawkins2004}
John~A Hawkins.
\newblock {\em {Efficiency and complexity in grammars}}.
\newblock Oxford Univ Press, Oxford, 2004.

\bibitem{Hawkins2007}
John~A. Hawkins.
\newblock {Processing typology and why psychologists need to know about it}.
\newblock {\em New Ideas in Psychology}, 25(2):87--107, August 2007.

\bibitem{Humboldt1972}
W.~von Humboldt.
\newblock {\em Linguistic Variability and Intellectual Development}.
\newblock University of Pennsylvania Press, Philadelphia, PAadelphia, 1972.

\bibitem{Hume2013}
Elizabeth Hume and Fr\'{e}d\'{e}ric Mailhot.
\newblock {The Role of Entropy and Surprisal in Phonologization and Language
  Change}.
\newblock In Alan C.~L. Yu, editor, {\em Origins of Sound Patterns: Approaches
  to Phonologization}, pages 29--50. Oxford University Press, Oxford, UK, 2013.

\bibitem{JaegerNorcliffe2009}
T~F Jaeger and E~J Norcliffe.
\newblock {The cross-linguistic study of sentence production}.
\newblock {\em Language and Linguistics Compass}, 3:1--22, 2009.

\bibitem{Jaeger2010}
T~Florian Jaeger.
\newblock {Redundancy and reduction: speakers manage syntactic information
  density}.
\newblock {\em Cognitive Psychology}, 61(1):23--62, August 2010.

\bibitem{Jel97}
Frederick Jelinek.
\newblock {\em Statistical Methods for Speech Recognition}.
\newblock MIT Press, Cambridge, MA, 1997.

\bibitem{Kirby2008}
Simon Kirby, Hannah Cornish, and Kenny Smith.
\newblock {Cumulative cultural evolution in the laboratory: an experimental
  approach to the origins of structure in human language.}
\newblock {\em Proceedings of the National Academy of Sciences of the United
  States of America}, 105(31):10681--6, August 2008.

\bibitem{Kirby2007}
Simon Kirby, Mike Dowman, and Thomas~L Griffiths.
\newblock Innateness and culture in the evolution of language.
\newblock {\em Proceedings of the National Academy of Sciences},
  104(12):5241--5245, 2007.

\bibitem{kneser95}
Reinhard Kneser and Hermann Ney.
\newblock Improved backing-off for m-gram language modeling.
\newblock In {\em International Conference on Acoustics, Speech, and Signal
  Processing (ICASSP)}, volume~1, pages 181--184, Detroit, MI, 1995. IEEE.

\bibitem{koehn2009}
Philipp Koehn.
\newblock {\em Statistical machine translation}.
\newblock Cambridge University Press, 2009.

\bibitem{Kohler1991}
K~J Kohler.
\newblock {The Phonetics/Phonology Issue in the Study of Articulatory
  Reduction}.
\newblock {\em Phonetica}, 48(2-4):180--192, 1991.

\bibitem{Levy2005}
Roger Levy.
\newblock {\em {Probabilistic Models of Word Order and Syntactic
  Discontinuity}}.
\newblock PhD thesis, Stanford University, 2005.

\bibitem{Levy2008}
Roger Levy.
\newblock {Expectation-based syntactic comprehension.}
\newblock {\em Cognition}, 106(3):1126--77, March 2008.

\bibitem{Lewis2006}
Richard~L Lewis, Shravan Vasishth, and Julie~a {Van Dyke}.
\newblock {Computational principles of working memory in sentence
  comprehension}.
\newblock {\em Trends in Cognitive Sciences}, 10(10):447--54, October 2006.

\bibitem{Lindblom1990}
Bj\"{o}rn Lindblom.
\newblock {Explaining phonetic variation: A sketch of the H\&H theory}.
\newblock In W~J Hardcastle and A~Marchal, editors, {\em Speech Production and
  Speech Modeling}, pages 403--439. Kluwer Academic Publishers, 1990.

\bibitem{Lohse2004}
Barbara Lohse, John~A Hawkins, and Thomas Wasow.
\newblock {Domain Minimization in English Verb-Particle Constructions}.
\newblock {\em Language}, 80(2):238--261, 2004.

\bibitem{Luce1998}
Paul~A Luce and David~B Pisoni.
\newblock {Recognizing Spoken Words: The Neighborhood Activation Model}.
\newblock {\em Ear and Hearing}, 19(1):1--36, 1998.

\bibitem{MacDonald2013}
Maryellen~C MacDonald.
\newblock {How language production shapes language form and comprehension}.
\newblock {\em Frontiers in Psychology}, 4(April):226, January 2013.

\bibitem{Magerman94}
David Magerman.
\newblock {\em Natural Language Parsing as Statistical Pattern Recognition}.
\newblock PhD thesis, Stanford University, 1994.

\bibitem{Magnuson2007b}
James~S Magnuson, James~A Dixon, Michael~K Tanenhaus, and Richard~N Aslin.
\newblock {The dynamics of lexical competition during spoken word recognition.}
\newblock {\em Cognitive science}, 31(1):133--56, February 2007.

\bibitem{Manin2006}
D~Yu Manin.
\newblock {Experiments on predictability of word in context and information
  rate in natural language}.
\newblock pages 1--12, 2006.

\bibitem{treebank}
Mitchell~P. Marcus, Beatrice Santorini, and Mary~Ann Marcinkiewicz.
\newblock Building a large annotated corpus of {E}nglish: {T}he {P}enn
  treebank.
\newblock {\em Computational Linguistics}, 19(2):313--330, June 1993.

\bibitem{McDonald2003}
Scott~A. McDonald and Richard~C. Shillcock.
\newblock {Low-level predictive inference in reading: the influence of
  transitional probabilities on eye movements}.
\newblock {\em Vision Research}, 43(16):1735--1751, July 2003.

\bibitem{Nowak2000}
Martin~A Nowak, Joshua~B Plotkin, and Vincent~AA Jansen.
\newblock The evolution of syntactic communication.
\newblock {\em Nature}, 404(6777):495--498, 2000.

\bibitem{Ohala1988}
John~J Ohala.
\newblock {Discussion of Bjoern Lindblom's 'Phonetic Invariance and the
  adaptive nature of speech'}.
\newblock In {\em Working Models of Human Perception}. Academic Press, London,
  UK, 1988.

\bibitem{Piantadosi2011}
Steven~T Piantadosi, Harry Tily, and Edward Gibson.
\newblock {Word lengths are optimized for efficient communication}.
\newblock {\em PNAS}, 108(9):3526 --3529, 2011.

\bibitem{Piantadosi2012}
Steven~T Piantadosi, Harry Tily, and Edward Gibson.
\newblock {The communicative function of ambiguity in language.}
\newblock {\em Cognition}, 122(3):280--91, March 2012.

\bibitem{Pierrehumbert2002}
Janet~B Pierrehumbert.
\newblock {Word-specific phonetics}.
\newblock In Carlos Gussenhoven and Natasha Warner, editors, {\em Laboratory
  phonology 7}, pages 101--139. Mouton de Gruyter, Berlin, 2002.

\bibitem{Pinker2005}
Steven Pinker and Ray Jackendoff.
\newblock {The faculty of language: what's special about it?}
\newblock {\em Cognition}, 95(2):201--36, March 2005.

\bibitem{Ros2015}
Idoia Ros, Mike Santesteban, Kumiko Fukumura, and Itziar Laka.
\newblock Aiming at shorter dependencies: the role of agreement morphology.
\newblock {\em Language, Cognition, and Neuroscience}, 2015.

\bibitem{Slobin1975}
Dan~I Slobin.
\newblock {Language Change in Childhood and in History}.
\newblock {\em Working Papers of the Language Behavior Research Laboratory},
  41:185--214, 1975.

\bibitem{Smith2013}
Nathaniel~J Smith and Roger Levy.
\newblock {The effect of word predictability on reading time is logarithmic.}
\newblock {\em Cognition}, 128(3):302--19, September 2013.

\bibitem{Szmrecsanyi2004}
Benedikt~M Szmrecs\'{a}nyi.
\newblock {On Operationalizing Syntactic Complexity}.
\newblock In {\em JADT 2004 : 7es Journ\'{e}es internationales d'Analyse
  statistique des Donn\'{e}es Textuelles}, pages 1031--1038, 2004.

\bibitem{Schijndel2015}
Marten van Schijndel and William Schuler.
\newblock Hierarchic syntax improves reading time prediction.
\newblock In {\em Proceedings of the 2013 Meeting of the North American chapter
  of the Association for Computational Linguistics (NAACL-15)}, 2015.

\bibitem{vanSon2003}
R~J J~H van Son and Louis C~W Pols.
\newblock {How efficient is speech?}
\newblock {\em Proceedings Institute of Phonetic Sciences, University of
  Amsterdam}, 25:171--184, 2003.

\bibitem{Vasishth2005}
Shravan Vasishth and R~L Lewis.
\newblock An activation-based model of sentence processing as skilled memory
  retrieval.
\newblock {\em Cognitive science}, 29(3):375--419, 2005.

\bibitem{Wasow2002}
Thomas Wasow.
\newblock {\em Post-verbal behavior}.
\newblock CSLI Publications, Stanford, CA, 2002.

\bibitem{Wedel2006}
Andrew Wedel.
\newblock {Exemplar models, evolution and language change}.
\newblock {\em The Linguistic Review}, 23:247--274, 2006.

\bibitem{Wedel2013a}
Andrew Wedel, Scott Jackson, and Abby Kaplan.
\newblock {Functional Load and the Lexicon: Evidence that Syntactic Category
  and Frequency Relationships in Minimal Lemma Pairs Predict the Loss of
  Phoneme contrasts in Language Change}.
\newblock {\em Language and Speech}, 56(3):395--417, July 2013.

\bibitem{ctb}
Nianwen Xue, Fei Xia, Fu-Dong Chiou, and Martha Palmer.
\newblock The penn chinese treebank: Phrase structure annotation of a large
  corpus.
\newblock {\em Natural Language Engineering}, 11:207--238, 2005.

\bibitem{Yamashita2001}
Hiroko Yamashita and Franklin Chang.
\newblock {"Long before short" preference in the production of a head-final
  language}.
\newblock {\em Cognition}, 81:45--55, 2001.

\bibitem{Zipf1949}
George~K. Zipf.
\newblock {\em {Human Behavior and the Principle of Least Effort}}.
\newblock Addison-Wesley, New York, 1949.

\end{thebibliography}

\end{document}